\documentclass{agujournal2019}
\usepackage{url} 
\usepackage{lineno}
\usepackage[inline]{trackchanges} 
\usepackage{soul}
\draftfalse

\journalname{relevant journal}

\begin{document}
\expandafter\def\expandafter\UrlBreaks\expandafter{\UrlBreaks
  \do\a\do\b\do\c\do\d\do\e\do\f\do\g\do\h\do\i\do\j%
  \do\k\do\l\do\m\do\n\do\o\do\p\do\q\do\r\do\s\do\t%
  \do\u\do\v\do\w\do\x\do\y\do\z\do\A\do\B\do\C\do\D%
  \do\E\do\F\do\G\do\H\do\I\do\J\do\K\do\L\do\M\do\N%
  \do\O\do\P\do\Q\do\R\do\S\do\T\do\U\do\V\do\W\do\X%
  \do\Y\do\Z}
%
%


\title{TPTNet: A Data-Driven Temperature Prediction Model Based on Turbulent Potential Temperature}

\authors{Jun Park\affil{1},Changhoon Lee\affil{1,2}}

\affiliation{1}{School of Mathematics and Computing, Yonsei University, Seoul 03722, Korea}
\affiliation{2}{School of Mechanical Engineering, Yonsei University, Seoul 03722, Korea}

\correspondingauthor{Changhoon Lee}{clee@yonsei.ac.kr}




\begin{keypoints}
\item Only 2m temperature data measured at the weather station is used as input for the prediction of temperature at finite forecast hours.
\item Turbulent fluctuation temperature relative to the climatological yearly and daily periodic variations with the altitude adjustment using potential temperature is considered in data-driven learning.
\item Up to 12 hours of forecast hour, TPTNet's prediction outperforms NWP with the less scattered errors over stations.
\end{keypoints}

%
%

%
%

\begin{abstract}
A data-driven model for predicting the surface temperature using neural networks was proposed to alleviate the computational burden of numerical weather prediction (NWP). Our model, named TPTNet uses only 2m temperature measured at the weather stations of the South Korean Peninsula as input to predict the local temperature at finite forecast hours. The turbulent fluctuation component of the temperature was extracted from the station measurements by separating the climatology component accounting for the yearly and daily variations. The effect of station altitude was then compensated by introducing a potential temperature. The resulting turbulent potential temperature data at irregularly distributed stations were used as input for predicting the turbulent potential temperature at forecast hours through three trained networks based on convolutional neural network (CNN), Swin Transformer, and a graphic neural network (GNN). The prediction performance of our network was compared with that of persistence and NWP, confirming that our model outperformed NWP for up to 12 forecast hours.
\end{abstract}

\section*{Plain Language Summary}
This study proposes a new prediction model based on deep learning to predict the temperature distribution using only the measured temperature at weather stations to alleviate the computational burden of numerical weather prediction (NWP). From 20 year data of South Korea, the mean periodic component accounting for yearly and daily variations was identified, and the fluctuations from this mean variation were considered for training the prediction network. The altitudes of the weather stations were also considered for training by introducing the potential temperature. The trained network was tested for unseen data for 2020, indicating that the model outperformed NWP and thus provided reliable temperature prediction for up to 12 h.

\section{Introduction}
Prior knowledge of the 2m temperature within a day has a significant impact on people's daily lives.  To predict temperature, weather forecasters rely on NWP, a model that numerically predicts weather by solving a proper physical model in the form of partial differential equations assisted by data assimilation. However, it requires enormous computational resources, such as the most advanced supercomputers, and a long computation time \cite{Pathak2022}. In particular, short-term predictions are difficult, because they are heavily influenced by data assimilation \cite{lam2022}. 

Recently, deep learning-based weather predictions have been attempted as a remedy to resolve the disadvantages of NWP, as reviewed in \citeA{Schultz2021}. In particular, reanalysis data have been used to train and test various neural network algorithms. \citeA{Rasp2021} made global weather predictions using a deep residual convolutional neural network (Resnet) on the WeatherBench data \cite{Rasp2020}. \citeA{Weyn2021} used the UNet-based on convolutional neural network to predict global weather at a resolution of 150 km. \citeA{Keisler2022,lam2022} made global weather predictions using ECMWF's ERA5 data \cite{Hersbach2020} with an autoregressive model adopting Graph Neural Network (GNN). \citeA{Gong2022} developed a static advertising video prediction model using the Generative Adversarial Network (GAN), Variational Auto Encoder, and ConvLSTM to predict weather in the central European area using the ERA5 reanalysis data set. \citeA{Hu2023} made global weather predictions by learning the WeatherBench data set with a variational recurrent neural network based on Swin Transformer blocks \cite{Liu_2021_ICCV}. Some neural-network-based prediction models outperform NWP. However, reanalysis data has been known to be less reliable than observation data directly measured at the weather stations \cite{mooney2011,Zhu2023}. Furthermore, the construction of the reanalysis data is dependent on NWP.


However, few studies have attempted to predict temperature through deep learning using observation data from weather stations \cite{Hou2022,Azari2022,Zhu2023}. \citeA{Hou2022} and \citeA{Azari2022} made temperature predictions at one weather station through time series modeling using various machine learning method such as LSTM.  \citeA{Zhu2023} considered for the first time the spatio-temporal modeling of temperature data at multiple weather stations. They created a dataset called Weather2K using weather data observed at 2,130 ground weather stations in China and trained a neural network based on the Multi Graph Convolution network. To the best of our knowledge, this study is the only attempt to predict temperature through neural networks using ground weather station data. However, the feasibility of this approach has not yet been fully explored.

From our experience with deep learning in various applications for the prediction, control, and modeling of turbulence \cite{Kim2020inflow, Kim2020heat, Kim2021sup, Kim2022les, Kim2023heat2, Lee2023drag}, we found that properly trained networks based on CNN or GAN can accurately predict turbulent fields of velocity and temperature. Recently, we developed a neural network based on a CNN or GAN to predict the vorticity field at finite lead times of up to half the integral time scale from an initial distribution in two-dimensional turbulence \cite{Kim2023twod}. Inspired by this finding, we developed a new neural network called TPTNet to predict the local 2m temperature distribution during forecast hours within a day using ground station data from the South Korean Peninsula.

Weather prediction in East Asia by using global NWP data is relatively difficult  \cite{lam2022,ramavajjala2023verification}. South Korea, located in East Asia, is an area with severe topographical changes ranging from 0m to 1800m above sea level in a small area of 100,210$km^2$ in four distinct seasons, resulting in wide variations in temperature throughout the country and large changes throughout the year. As of 2020, approximately 380 ground weather stations managed by the Korea Meteorological Administration are distributed over the peninsula. Although the number density of stations of one per 250$km^2$ is relatively higher than that per 2,800$km^2$ of China \cite{Zhu2023} and one per 894$km^2$ of the United States, the Peninsula is surrounded by sea and North Korea, for which the availability of observation data is limited, which makes prediction based on ground station data challenging. In this study, we propose a methodology for deep learning to predict the local 2m temperature distribution during forecast hours within a day in the South Korean Peninsula using only ground weather station data. The performance of the trained networks was compared with NWP. 

The remainder of this paper is organized as follows. Section 2 explains the methodology of deep learning with data preprocessing. Detailed training data and their statistical features are provided in Section 3. The training and testing results of the proposed network models are discussed in Section 4. Finally, we conclude out study in Section 4.

\section{Methods}
\subsection{Data preprocessing}
\begin{figure}
\noindent\includegraphics[width=\textwidth]{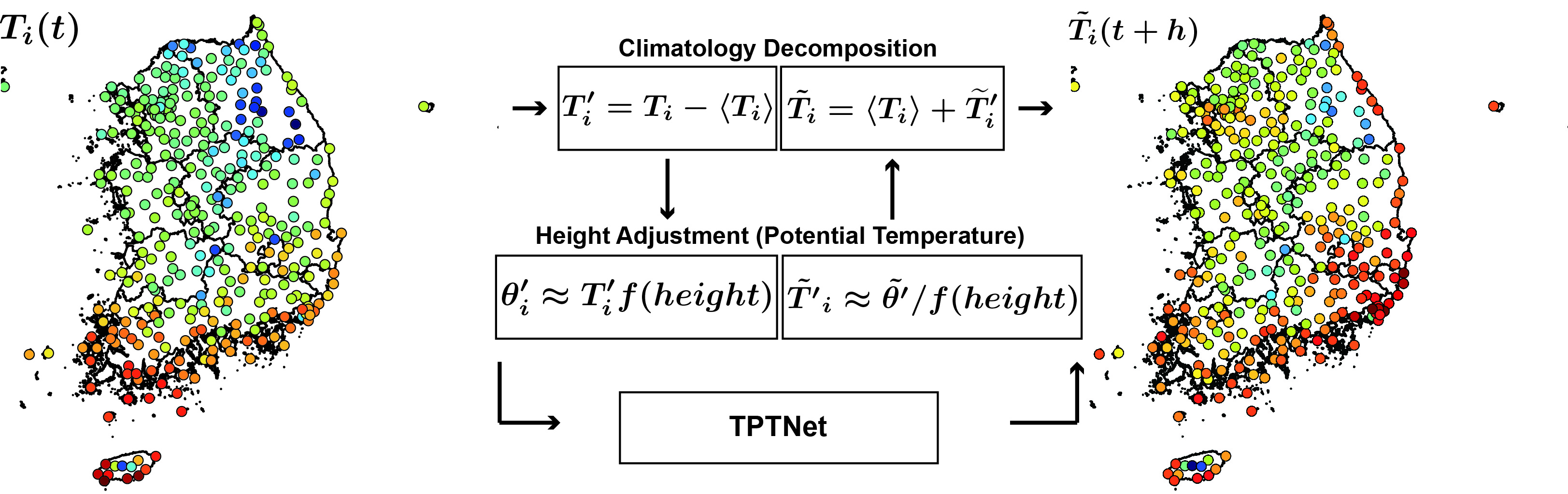}
\caption{
Schematic of temperature forecasting workflow. 2m temperature at station ($T_i$) is converted to turbulent potential temperature ($\theta’_i$) through the climatology decomposition and height adjustment using potential temperature. Temperature ($T_i$) is divided into the yearly and daily periodic data ($\langle{T_i}\rangle$) and turbulent temperature ($T_i’$). $T'_i$ is converted to $\theta'_i$ using Equation \ref{eq:tpt}. Turbulent potential temperature data at time $t$ are injected into TPTNet as input. TPTNet predicts the turbulent potential temperatures at time $t+h$. Finally, $\widetilde{T}_i$ at $t+h$ is produced through height adjustment and mean decomposition.}
\label{fig:schematic}
\end{figure}

2m temperature data $T_i( t)$ at a station located at $(x_i, y_i)$ used for training and testing the proposed networks were provided by the Korea Meteorological Administration. Here, $i$ denotes the ID number of a station and $i=1, \cdots, 366$. Temperature data for 20 years from 2000 to 2019 at one minute resolution were used for training.  The average distance between a station and its nearest station is approximately 13 $km$. The altitude above sea level at the stations ranged from 5 $m$ to 968 $m$, with an average of 124 $m$.

Before training, we performed two types of data preprocessing for efficient training. First, the temperature was decomposed into the climatological mean and fluctuation parts:
 \begin{equation}
     T_i (t) = \langle{T_i\rangle} (t) + T_i'(t),
     \label{eq:T}
 \end{equation}
where $\langle{T_i\rangle}$ is the expected periodic part accounting for the annual and daily variations and $T_i'$ denotes the deviation from $\langle{T_i\rangle}$, which is called the turbulent part. The method for obtaining $\langle{T_i\rangle}$ is explained below.

Next, we introduced the potential temperature to consider the effect of station altitude. This was necessary because the potential temperature is a conserved quantity similar to vorticity in two-dimensional flows, and the altitude of ground weather stations in Korea varies widely. The potential temperature at the $i$th station $\theta_i$ is defined as:
\begin{equation}
   \theta_i = T_i \left(\frac{P_0}{P_i}\right)^{R/C_p} \simeq T_i f(z_i) ,~~\theta'_i = T'_i f(z_i)
   \label{eq:tpt}
\end{equation}
where $P_0 (=1000 hPa)$ is the reference pressure at sea level and $P_i$ is the pressure at the $i$th station. $R$ and $C_p$ are the gas constants of air and specific heat, respectively. $z_i$ is the altitude of the $i$-th station. Using the pressure measured at 63 stations for 20 years (2000 $\sim$ 2019), $f(z)$ was approximately obtained, as described in the Supporting Information (Text S1, Figure S1). Therefore, the fluctuation part $\theta'_i$ was used for training and is referred to as the turbulent potential temperature (TPT). Our network, called TPTNet predicts the potential temperature distribution during forecast hours; a schematic of our network is provided in Figure \ref{fig:schematic}.

For the training of the TPTNet, we chose three deep learning models: CNN, Swin Transformer, and GNN. The training of the CNN and Swin Transformer requires data on a uniform mesh. The mesh data were generated from irregularly distributed station data through spatial interpolation using the ordinary kriging method \cite{Isaaks1989}, incorporating the UTM coordinates of the stations. The results of the comparison of the various spatial interpolation methods are provided in the Supporting Information (Text S2, Table S1). From the predicted data on the meshes, station predictions were obtained using bilinear interpolation. 

\subsection{Evaluation metrics}
The performance of the developed network was assessed by comparing the predicted and measured temperatures at each station. At each station, the corresponding evaluation metric is defined as
\begin{equation}
    RMSE_i=\sqrt{\frac{1}{ N_t} \sum_t^{N_t} {(\langle{T_i\rangle}(t)+\widetilde{T'_i}(t)-T_i(t))^2}}
\end{equation}
where $\widetilde{T'_i}$ is the turbulent temperature predicted by the networks or NWP and $N_t$ is the number of tests. To compare the prediction errors between the different prediction models, the average error was considered.
\begin{equation}
    RMSE=\frac{1}{ N_s} \sum_{i=1}^{N_s} RMSE_i ,
    \label{eq:RMSE}
\end{equation}
where $N_s$ is the number of stations.

To assess networks based on the CNN, the root-mean-squared error of the meshes was used.
\begin{equation}
    RMSE_{m} = \frac{1}{N_{inland}} \sum_{x,y \in land}^{N_{inland}} \sqrt{\frac{1}{N_t} \sum_t^{N_t} {(\widetilde{\theta'}_{m}(x,y,t)-\theta'_{m}(x,y,t))^2}}
    \label{eq:RMSEm}
\end{equation}
where $\widetilde{\theta'}_{m}$ and $\theta'_{m}$ are the predicted and interpolated measured turbulent potential temperatures at the mesh points, respectively. $N_{inland}$ represents the number of grid points on land.

\subsection{NWP data}
The baseline model against which the performance of the proposed network was compared was NWP. We adopted the NWP results provided by Korea's Local Data Assimilation and Prediction System (LDAPS). We use the 2020 version of the LDAPS as a comparison model \cite{kmaAlmanac}. The base model of LDAPS is the United model of the United Kingdom. It performed 48-hour prediction of the meteorological variables on 1,780km $\times$ 1,720km at a resolution of 1.5km on the horizontal plane. Table \ref{tab:LDAPS} provides information on the LDAPS considered in this study. 2m temperature data at the station location were obtained using a bilinear interpolation from the prediction results at four nearby grid points and compared with the observation data and the prediction by our network for the year 2020. Five forecast hours (1, 3, 6, 12, and 24 h) were considered for comparison.

\begin{table}
 \caption{\label{tab:LDAPS} Composition of LDAPS in 2020}
\centering
\begin{tabular}{l l} 
 \hline
 horizontal grid size/grid dimension & 1.5 km / 1188 x 1148\\
 vertial layers & 70 ($\sim$40 km) \\
 data assimilation & 3DVAR (FGAT, IAU) \\
 time step & 60 s\\
 prediction hour & 48 h\\
 dynamics cores & ENDGame \\
 planetary boundary layer & Revised entrainment fluxes plus new scalar flux-gradient option \\
 radiative process & Edwards-Slingo spectral band radiation \\
 cloud physics & Mixed-phase scheme with graupel \\
\hline
\end{tabular}
\end{table}



\subsection{TPTNet}
\begin{figure}
\noindent\includegraphics[width=\textwidth]{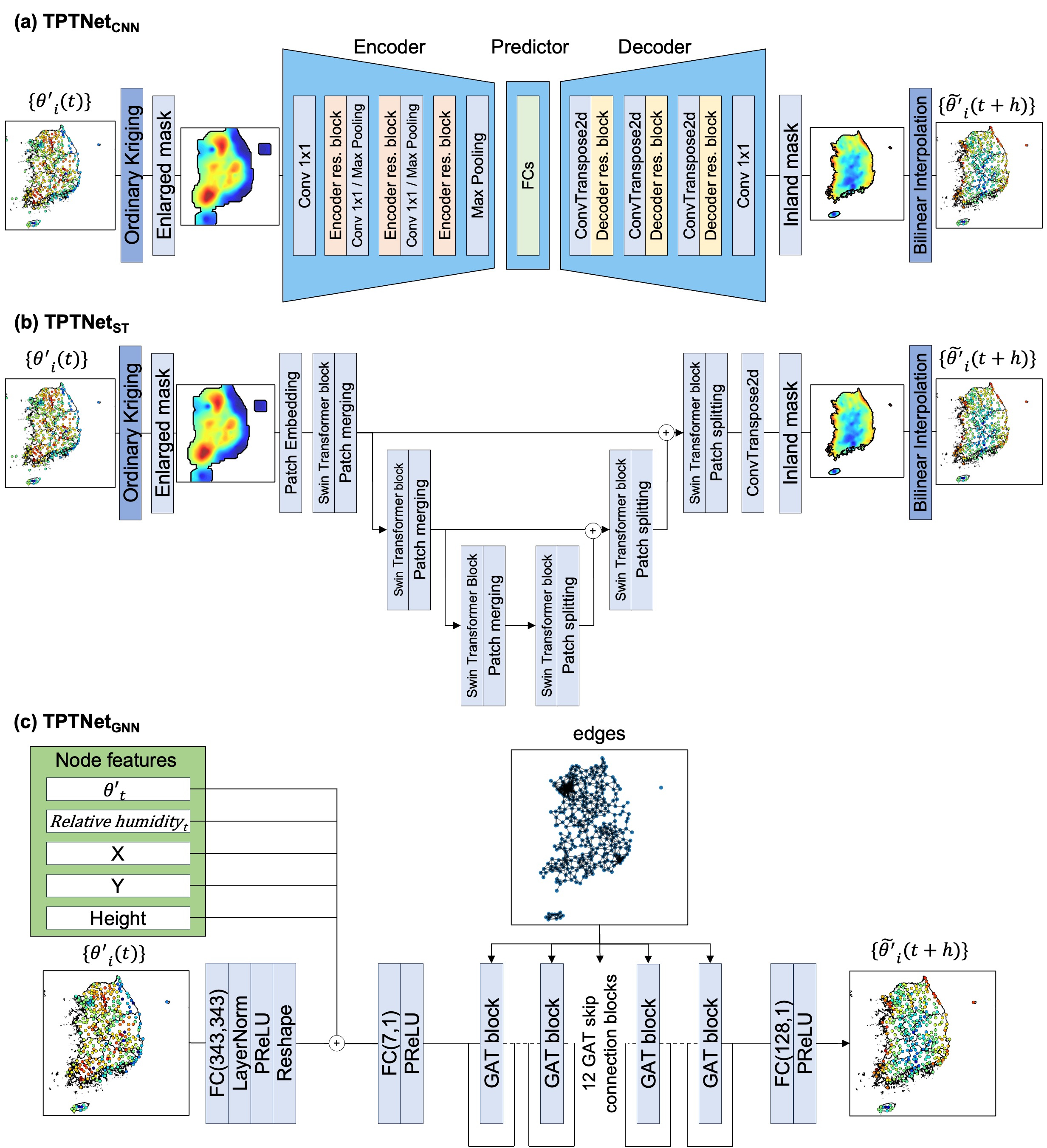}
\caption{Architecture of (a) $TPTNet_{CNN}$, (b) $TPTNet_{ST}$ and (c) $TPTNet_{GNN}$}
\label{fig:architecture}
\end{figure}

To train the TPTNet, we tested three popular deep-learning models: CNN, Swin Transformer, and GNN. These models are referred to as $TPTNet_{CNN}, TPTNet_{ST}$ and $TPTNet_{GNN}$, respectively, for which the architectures are illustrated in Figure \ref{fig:architecture}. For $TPTNet_{CNN}$ and $TPTNet_{ST}$, the station data were transformed into mesh data through ordinary kriging, whereas no such transformation was necessary for $TPTNet_{GNN}$, which works for irregularly distributed data. 

As shown in Figure \ref{fig:architecture}(a), $TPTNet_{CNN}$ is an AutoEncoder-based network composed of encoders and decoders, and the size of the latent vector is set to 128. The encoder and decoder were designed as residual (res) block-based structures. Batch normalization and pixelwise norms were used in the decoder. We used pixelwise norm presented in \citeA{karras2017}. A predictor exists between the encoder and the decoder that converts the latent vector into a latent vector at the target time. A fully connected layer and the LeakyReLU activation function were used as predictors. The number after Conv is the filter size for the 2D convolution.

The Swin Transformer architecture introduced by \citeA{liu2022swin} constitutes $TPTNet_{ST}$. As shown in Figure \ref{fig:architecture}(b), we adopted six Swin Transformer layers, with the initial three layers functioning as encoders. Notably, we employed patch merging to reduce the size effectively, followed by three subsequent 3 layers that served as decoders. To counterbalance this reduction, we leverage patch splitting to increase the size. The encoder and decoder modules were integrated using a U-Net structure. In terms of the configuration, we delineated the depth of Swin Transformer layers as [2, 4, 12, 12, 4, 2], imparting distinct characteristics to the different layers. In addition, the allocation of attention heads varies across layers, with counts specified as [3, 6, 12, 12, 6, 3]. We set the window size to eight while maintaining a consistent patch size of eight. The patch-embedding dimensions are defined as 96.

For $TPTNet_{CNN}$ and $TPTNet_{ST}$, ordinary kriging was performed to generate 128 × 128 full-mesh data points. To make predictions within an integral timescale on land, data over the sea, far from land, are not necessary. Therefore, an enlarged mask used for prediction was placed in the input stage by referring only to the grids within 30 km of the land and setting the other areas at 0. To specifically predict $\widetilde{\theta'}$ on the land, we used an inland mask during the prediction process. Subsequently, when evaluating  performance of the model, we considered the loss exclusively based on $\widetilde{\theta'}$ for land areas only. The corresponding MSE loss functions ($L_{TPTNet}$) for $TPTNet_{CNN}$ and $TPTNet_{ST}$ are
\begin{equation}
    L_{TPTNet} =  \frac{1}{N_{inland}} \sum_{x,y \in land } w(x,y) (\widetilde{\theta'}_{m}(x,y)- \theta'_{m}(x,y))^2 ,
\end{equation}
where $w(x,y)$ is introduced to reflect the nearness of the mesh point to the station location because the eventual goal is to predict the temperature at the stations. Several methods to determine $w(x,y)$ depending on the station density were considered. However, these methods did not make meaningful improvements over the model without such considerations. Therefore, $w(x,y)=1$ for all the mesh points in our networks.


For $TPTNet_{CNN}$ and $TPTNet_{ST}$, errors due to kriging and interpolation in the preprocessing and postprocessing stages are inevitable because all the training is performed using the mesh data. To eliminate these errors, $TPTNet_{GNN}$ is constructed based on a graph-attention network (GAT) \cite{velivckovic2017graph}, as illustrated in Figure \ref{fig:architecture}(c). Graph edges were defined by connecting stations within a 30 km radius. Various experiments were conducted, including training the model using graphs generated based on the correlation between $\theta'_i$ values from each station as edges, connecting edges based on distance, and creating graphs divided into multiple segments based on distance intervals. Despite these efforts, it was observed that constructing graphs by connecting observation points within a 30km radius resulted in the best predictions. Additionally, it utilizes information such as relative humidity, x-y coordinates, and height as part of the node feature information for $TPTNet_{GNN}$. Therefore, the MSE loss function for $TPTNet_{GNN}$ is:
\begin{equation}
    L_{TPTNet} =  \frac{1}{N_{s}} \sum_{i=1}^{N_s}  (\widetilde{\theta'}_{i}- \theta'_{i})^2 .
\end{equation}
During training, the Adam optimizer was used, and beta1 was set to 0.9 and beta2 to 0.999. The learning rate is 1e-4. Because the training results did not change significantly even after learning for more than 100 epochs, TPTNet was tested after training for 100 epochs.


\section{Training Data}
\subsection{Data preprocessing}
Prior to generating data for training, raw data of 2m temperature measured at the ground stations underwent a series of preprocessing procedures: a quality control (QC) test and extraction of the climatology part $\langle{T_i\rangle}$. First, to enhance data quality, a QC test was conducted \cite{WMO-Guidelines}. For 20 years (2000 $\sim$ 2019), the lowest and highest temperatures measured in Korea were -32.6 $^\circ C$ and 41 $^\circ C$, respectively. Therefore, any data outside this range were classified as false. In addition, a turbulent temperature $T_i'$  deviating from $\langle{T_i\rangle}$ by greater than 20 K was classified as a false value. These false and null data were replaced with linearly interpolated data from the nearest available data. When the first or last data were false or null, extrapolation was performed by averaging nearby data. The amount of replaced data is less than 1 \%.

\begin{figure}
\noindent\includegraphics[width=\textwidth]{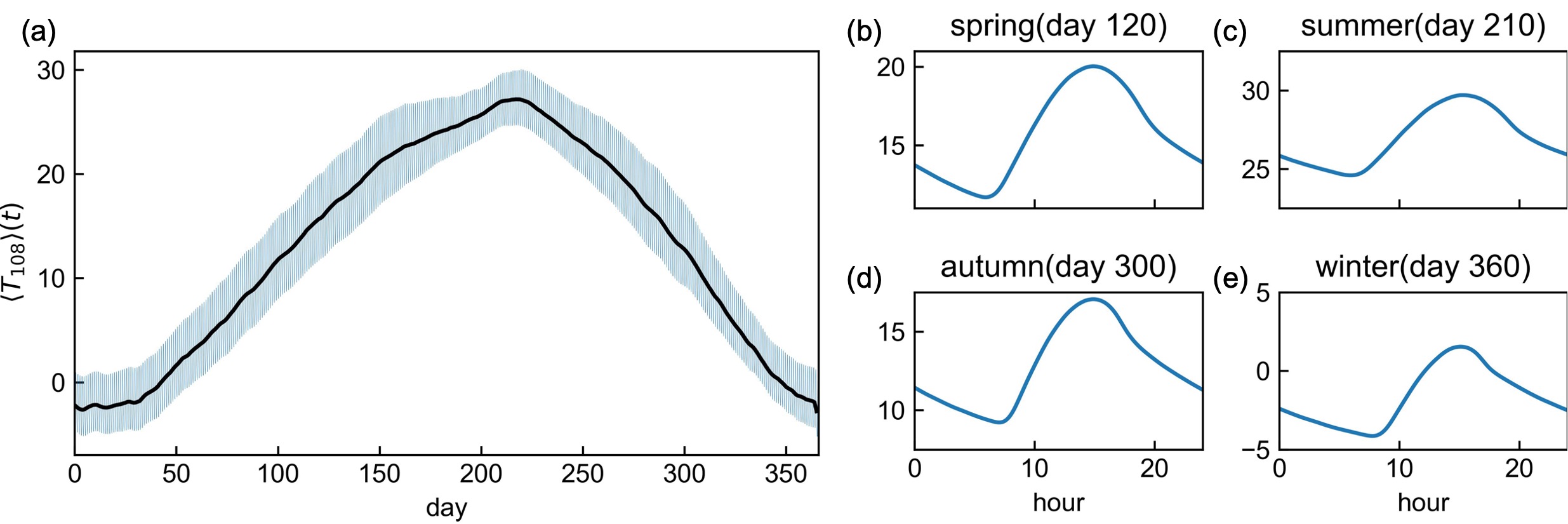}
\caption{An example distribution of the climatology accounting for the averaged yearly and daily variation of temperature at one station in Seoul $\langle T_{108} \rangle (t)$ (station ID is 108): (a) yearly distribution of temperature over 366 days and daily distribution over 24 hours in a chosen day of (b) Spring, (c) Summer, (d) Fall, and (e) Winter. Black line in (a) stands for daily averaged temperature. The time resolution is 1 minute. }
\label{fig:periodicity}
\end{figure}

The climatological temperature accounting for the yearly and daily variations $\langle{T_i\rangle}(t)$ was extracted by averaging $T_i(t)$ over a 20 year period. To smoothen the data, extra local averaging over approximately 21 days was performed. If one-minute resolution temperature data are expressed by $T_i(Y:D:H:M)$, 
\begin{equation}
 \langle{ T_i \rangle} (D:H:M) = \frac{1}{20 \cdot 21} \sum_{Y=2000}^{2019} \sum_{d=-10}^{10} T_i(Y:D+d:H:M)
\end{equation}
where $Y, D, H, M$ denote year, day, hour, and minute, respectively. The range of $D$ is $1 \sim 366$ to consider a leap year. An example $\langle{ T_i \rangle}$ for station ID 108 (Seoul) is shown in Figure \ref{fig:periodicity}. The local daily variations on Days 120, 210, 300, and 360 are also shown. Using this climatology temperature, persistence and climatology forecasts were readily made using $\tilde{T'}_i(t+h) = T'_i(t)$ and $\tilde{T'}_i(t+h)=0$, respectively.

To train $TPTNet_{CNN}$ and $TPTNet_{ST}$, the mesh temperature data were required. Before transforming the station data into mesh data, the turbulent temperature data $T_i'$ were converted to turbulent potential temperature data $\theta_i'$ by using Equation ~(\ref{eq:tpt}). Subsequently, the mesh data $\theta_{m}'$ are created by interpolation using an ordinary kriging method. The latitude and longitude ranges considered in this study were $[124.839 ^\circ E \sim 131.945 ^\circ E][33.041 ^\circ N \sim 38.710 ^\circ N ]$. The latitude and longitude coordinates were projected onto the EPSG5179 UTM coordinate system, which is an orthogonal system. Spatial interpolation was performed using the UTM coordinate system. The generated mesh had a resolution of 5 km and consisted of 128 $\times$ 128 grids covering 640km $\times$ 640km. This resolution was determined based on our conjecture that because the average distance between one station and the nearest station is 13km, data on the grids with 5km resolution can properly reflect the station data without waste.

To assess the error resulting from spatial interpolation, we retrieved the temperatures at the station locations from the generated mesh data and compared them with the original data. For comparison, we examined the interpolated temperatures obtained using the bilinear interpolation method. The average RMSE of the temperature was 0.5 K. 

To train our models, we randomly divided the mesh data collected from 2000 to 2019 into three sets: 80\% for training, 10\% for validation, and 10\% for interleaved testing. To test for unseen data, data for the year 2020 were chosen, and similar preprocessing was conducted to generate hourly data for the entire year. To facilitate learning, the data were normalized using standardization, which ensured a mean of 0 and a standard deviation of 1.

\subsection{Statistics of training data}
\begin{figure}
\noindent\includegraphics[width=\textwidth]{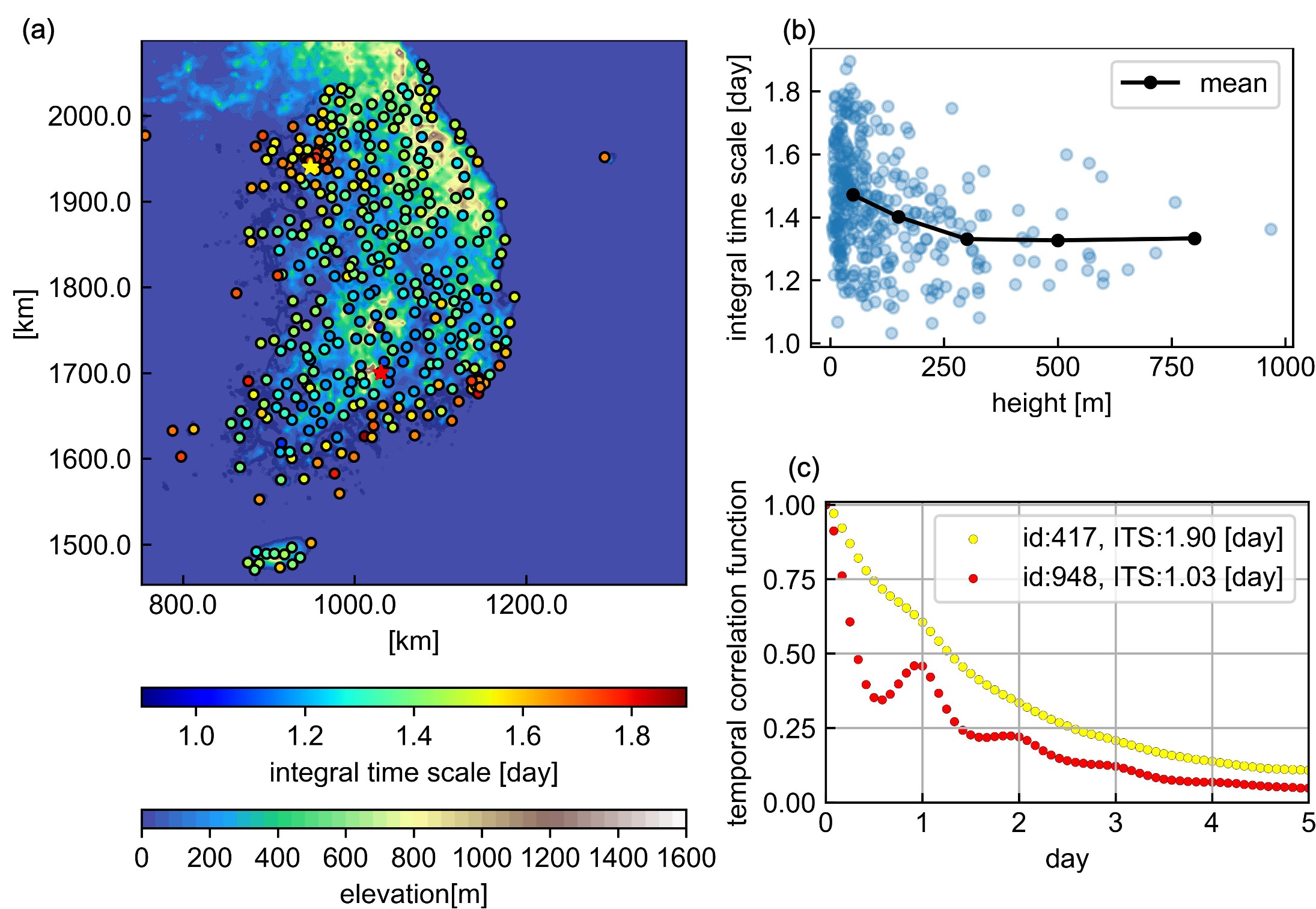}
\caption{Statistics of ITS of training data set. (a) ITS calculated at stations. The color of dots represents ITS at station position. The average is 1.433 day. Yellow and red stars indicate the stations with ID 417 and 948, each exhibiting the longest (1.9 day) and shortest ITS (1.03 day), respectively.  (b) ITS versus the altitude of stations. (c) Temporal correlation function for stations marked by yellow and red stars.}
\label{fig:its_compare}
\end{figure}

Before training the network models for prediction, statistical analysis was conducted using the turbulent temperature at station $T'_i(t)$. A key statistical quantity is the integral time scale (ITS), which represents the duration of information retention. ITS is defined by
\begin{equation}
ITS = \int_0^{\infty} \rho_i(s)~ds ,
\end{equation}
where $\rho_i(s)$ is the temporal correlation function defined as
\begin{equation}
\rho_i(s) = \frac{\overline{T'_i(t) T'_i(t+s)}}{\overline{T_i'^2}} ,
\end{equation}
where the overline denotes average over time. The ITS for each station is demonstrated in Figure \ref{fig:its_compare}(a), and the average ITS over all stations is 1.43 day (34.3 hours). The ITS ranged from 1.1 $\sim$ 1.8 days and the ITS at higher stations tended to be slightly shorter, as shown in Figure \ref{fig:its_compare}(b). The temporal correlation function at two stations (ID 417 and 948 marked by yellow and red stars in Figure \ref{fig:its_compare}(a)) shown in Figure \ref{fig:its_compare}(c) clearly indicates that, depending on the station location, the ITS can differ widely. It is also noteworthy that even after the decomposition of the periodic mean, the daily variation was not completely eliminated. Given that the average ITS is approximately 34 h, an attempt to predict the temperature during forecast hours of up to 24 h seems reasonable.

\begin{figure}
\noindent\includegraphics[width=\textwidth]{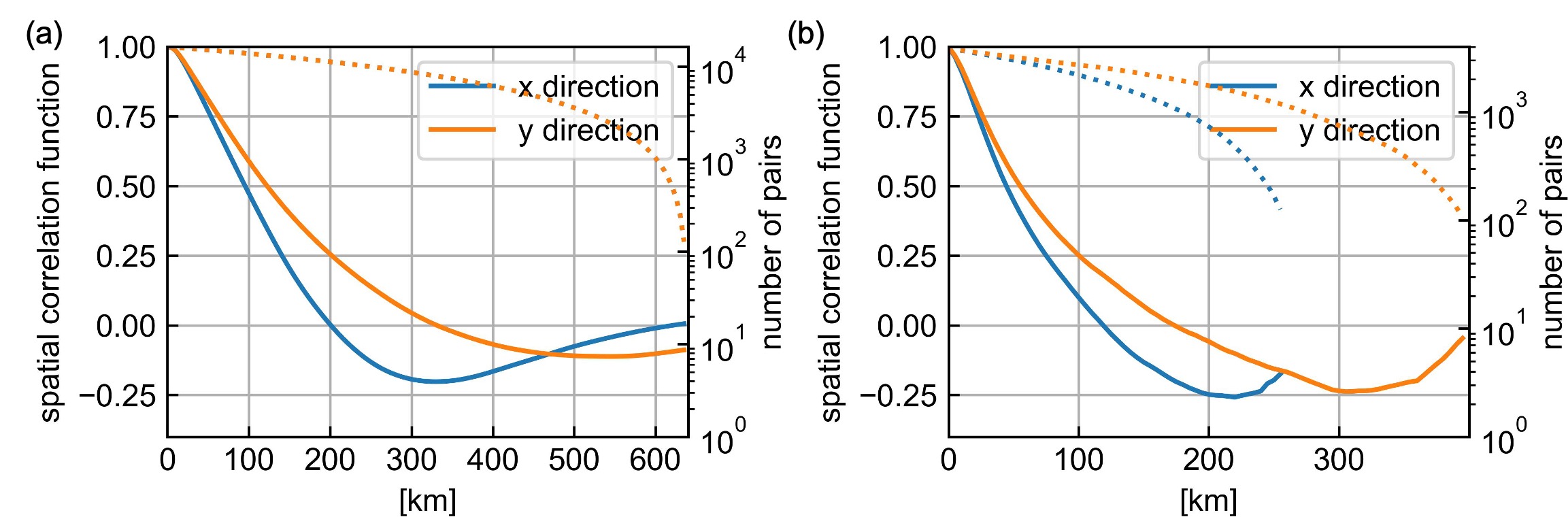}
\caption{Spatial correlation function obtained over (a) entire area including sea and (b) inland area. The blue line indicates the longitudinal correlation, while the orange line indicates the latitudinal correlation function. The solid line is the spatial correlation function and the dotted line is the number of pairs. Only those with 100 or more pair data are displayed.}
\label{fig:scfs}
\end{figure}

To assess the spatial correlation of the generated mesh data $\theta'_m$, we investigated the spatial correlation functions $R_h(r)$ and $R_v(r)$ in the longitudinal and latitudinal directions, respectively, which are defined as follows:
 \begin{eqnarray}
    R_h(r) & = & \frac{1}{N_t} \sum_t^{N_t} \frac{ \langle(\theta'_m(x,y,t)-\langle \theta'_m \rangle)(\theta'_m(x+r,y,t)-\langle \theta'_m \rangle) \rangle }
{\sigma(\theta'_m)^2 } \\
    R_v(r) & = & \frac{1}{N_t} \sum_t^{N_t} \frac{ \langle(\theta'_m(x,y,t)-\langle \theta'_m \rangle)(\theta'_m(x,y+r,t)-\langle \theta'_m \rangle) \rangle }
{\sigma(\theta'_m)^2 }
 \end{eqnarray}
where $\langle\rangle$ denotes averaging over the mesh points and $\sigma(\theta'_m)$ is the standard deviation in space. Figures \ref{fig:scfs}(a) and \ref{fig:scfs}(b) illustrate the spatial correlations obtained over the entire area, including the sea and inland regions. The number of grid-point pairs used to calculate the spatial correlation function is represented by the dotted line. We considered only the statistics obtained from more than 100 pairs of data.  The spatial correlation function provides an indication of how much temperature information at a specific grid point is correlated with that at neighboring points. Because ordinary kriging was performed based on the station data on land, the spatial correlation functions exhibited a longer correlation over the entire area, including the sea, than over the land area only because of the smoother distribution of interpolated temperature over the sea. Another noticeable observation is that the temperature is more correlated in the latitudinal direction than in the longitudinal direction. This is probably caused by the fact that mountain ridges are more aligned in the latitudinal direction than in the longitudinal direction in Korea; thus, temperature varies more inconsistently in the longitudinal direction. Overall, temperatures at the two points separated by 50 km were relatively well correlated, maintaining a correlation higher than 0.5. This observation supports the use of a CNN or Swin Transformer in the extraction of the spatial pattern of temperature through training.

Some grid points on the land were located at the boundary adjacent to the coast. Information at these points is at risk of loss during the multiple convolution processes in the network. To prevent such information loss, we extended the training data by including six additional grids on the sea (approximately 30 km) adjacent to the coast based on the spatial correlation function. This approach ensured that valuable coastal information was retained during the training of TPTNet.

\section{Results and Discussions}
\subsection{Optimization of $TPTNet$}
\begin{figure}
\noindent\includegraphics[width=\textwidth]{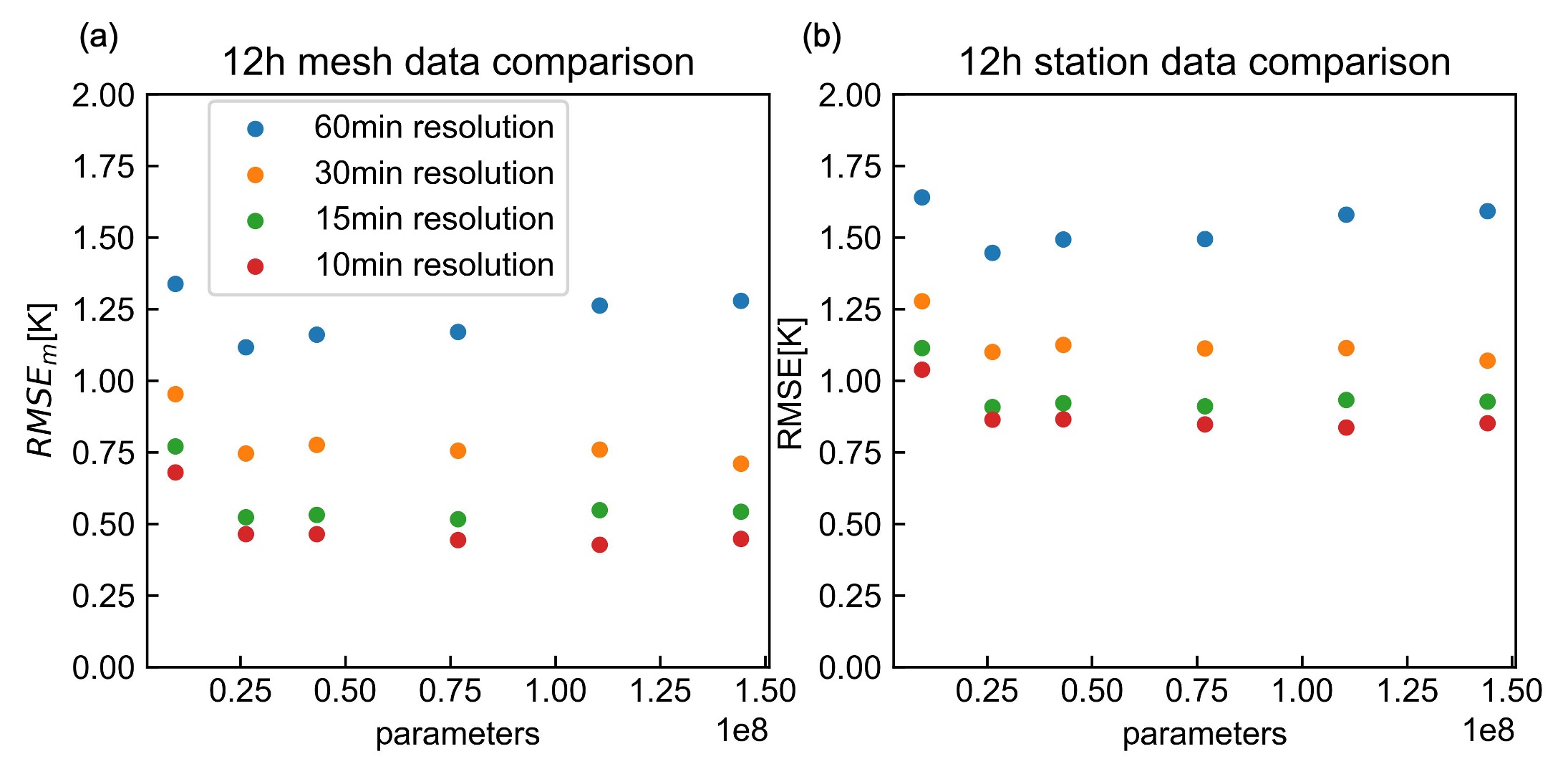}
\caption{The results of parameter dependence test of $TPTNet_{CNN}$ after training for 100 epochs. (a) errors on the meshes, $RMSE_m$ defined by Equation (\ref{eq:RMSEm}) and (b) errors on the stations, $RMSE$ defined by Equation (\ref{eq:RMSE}). Target prediction hour is 12 hours. } 
\label{fig:parameter_dep_test}
\end{figure}

For parameter optimization of the networks, we conducted a parameter dependence test with mainly  $TPTNet_{CNN}$, specifically targeting the 12-hour prediction. In particular, we investigated the effect of the size of the learning parameters and the amount of training data on the performance. The test results are shown in Figure \ref{fig:parameter_dep_test}.  When changing the sizes of the learning parameters, we focused on modifying the feature size of the FCs in the Predictor component, as shown in Figure \ref{fig:architecture}(a). By adjusting the intermediate feature (k in Figure S2) to 128, 65536, 131072, 262144, 393216, and 524288 in the FCs, we observed that the RMSE behaved non-monotonically. Interesting behavior of the RMSE was revealed in this parameter dependence test. When the amount of training data is sufficiently large such that data resolution is 30 minutes or below, as the number of training parameters increases, RMSE first decreases to reach a local minimum and then increases a little bit followed by decrease again as shown in Figure \ref{fig:parameter_dep_test}, which has been known as 'double descent phenomena' \cite{Belkin2019,nakkiran2021deep}. A similar behavior has also been observed in deep-learning-based weather forecasting \cite{Rasp2021}. From the behavior of RMSE for 10 min resolution data, we selected the optimum number of trainable parameters of $TPTNet_{CNN}$ as 110M, exhibiting the second local minimum of errors.

Furthermore, we investigated the effect of the amount of training data on the performance. This was achieved by increasing the size of the training dataset and reducing the temporal resolution of the data. As shown in Figure \ref{fig:parameter_dep_test}, as the resolution decreased from 60 min to 10 min, RMSE decreased but tended to converge to a certain value owing to the finite diversity of the data. In all tests hereinafter, 10 min resolution data were used for training. The number of trainable parameters of $TPTNet_{ST}$ and $TPTNet_{GNN}$ were 112M and 370K, respectively. The relatively smaller size of $TPTNet_{GNN}$ is due to the smaller number of stations (366) compared to the meshes required for $TPTNet_{CNN}$ and $TPTNet_{ST}$ (128 $\times$ 128). These sizes of trainable parameters were selected because any further increase in size did not significantly improve the performance. The wall-clock times required to learn one epoch on a single GPU machine (Nvidia GeForce RTX 3090) for $TPTNet_{CNN}, TPTNet_{ST}$ and $TPTNet_{GNN}$ were 20, 18, and 25 min, respectively, for the same amount of training data (10 min resolution). For complete training, 100 epochs were required for the three models, and the required times were 33, 30, and 41 h for $TPTNet_{CNN}$, $TPTNet_{ST}$ and $TPTNet_{GNN}$, respectively. From the comparison of RMSE between the mesh data and station data in the training of $TPTNet_{CNN}$, it is noticeable that RMSE of the station data is larger than RMSE of the mesh data by 0.4K on average, which is caused by the ordinary kriging and interpolation necessary for the transformation of data between meshes and stations.

\subsection{Interleaved test}
\begin{figure}
\noindent\includegraphics[width=\textwidth]{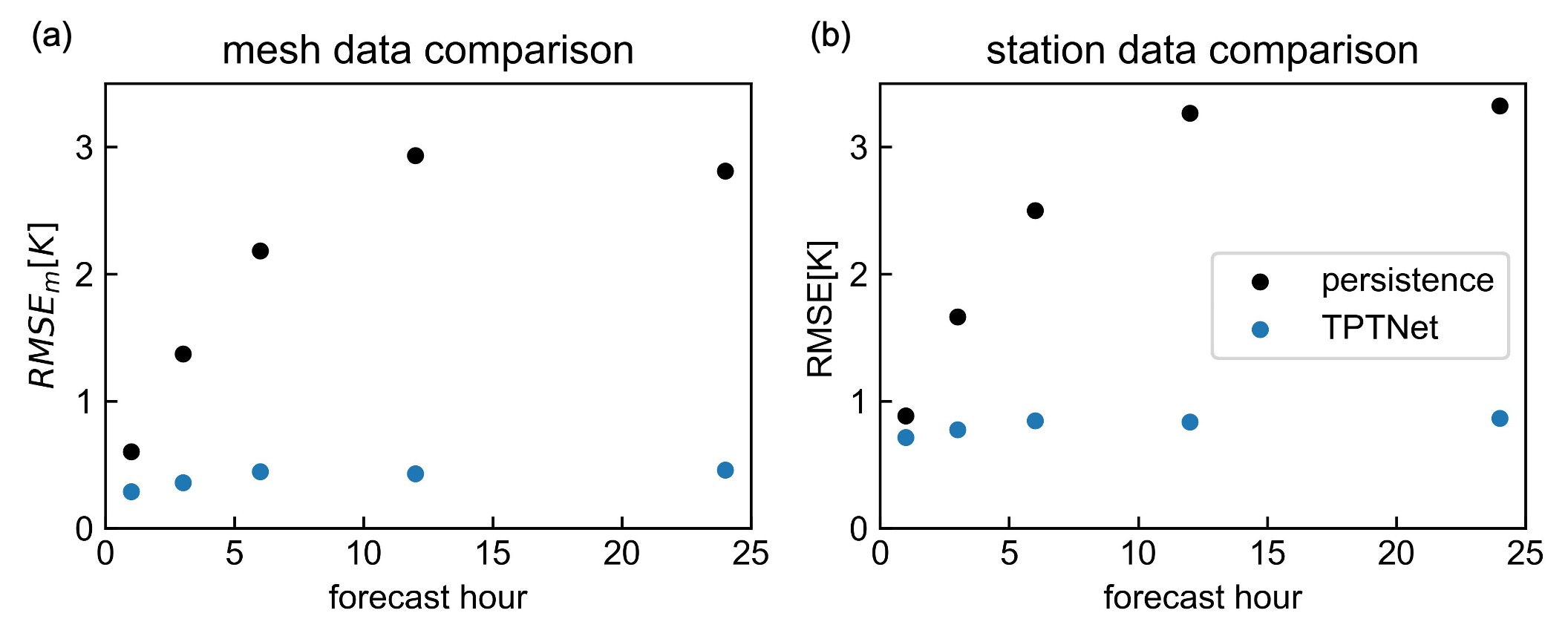}
\caption{The results of the interleaved test of $TPTNet_{CNN}$. (a) prediction errors on the mesh data and (b) errors on the station data for 5 forecast hours: 1, 3, 6, 12, and 24 hours.}
\label{fig:forecast_h_dep}
\end{figure}

Before testing the trained networks for the unseen data, we investigated the performance of the interleaved tests. Among the training data for 20 years (2000 $\sim$ 2019), randomly selected 10 \% data were used for the interleaved test. RMSE of the prediction using $TPTNet_{CNN}$ at five  forecast hours (1, 3, 6, 12, and 24 hours) is shown in  Figure \ref{fig:forecast_h_dep}. The model is trained separately for each forecast hour. We attempted rollout prediction by iteratively applying the model trained for a short time to a long time prediction, but the rollout prediction underperformed the single prediction. RMSEs for the mesh and station data were compared with those of the persistence model. Here, the persistence prediction is defined as $\tilde{\theta'}_m(t+h) = \theta'_m(t)$. RMSE of $TPTNet_{CNN}$ for the mesh data is below 0.5K for all lead times, while RMSE for the station data remains below 0.9K, exhibiting quite an insensitive dependence on the forecast hour.

\begin{figure}
\noindent\includegraphics[width=\textwidth]{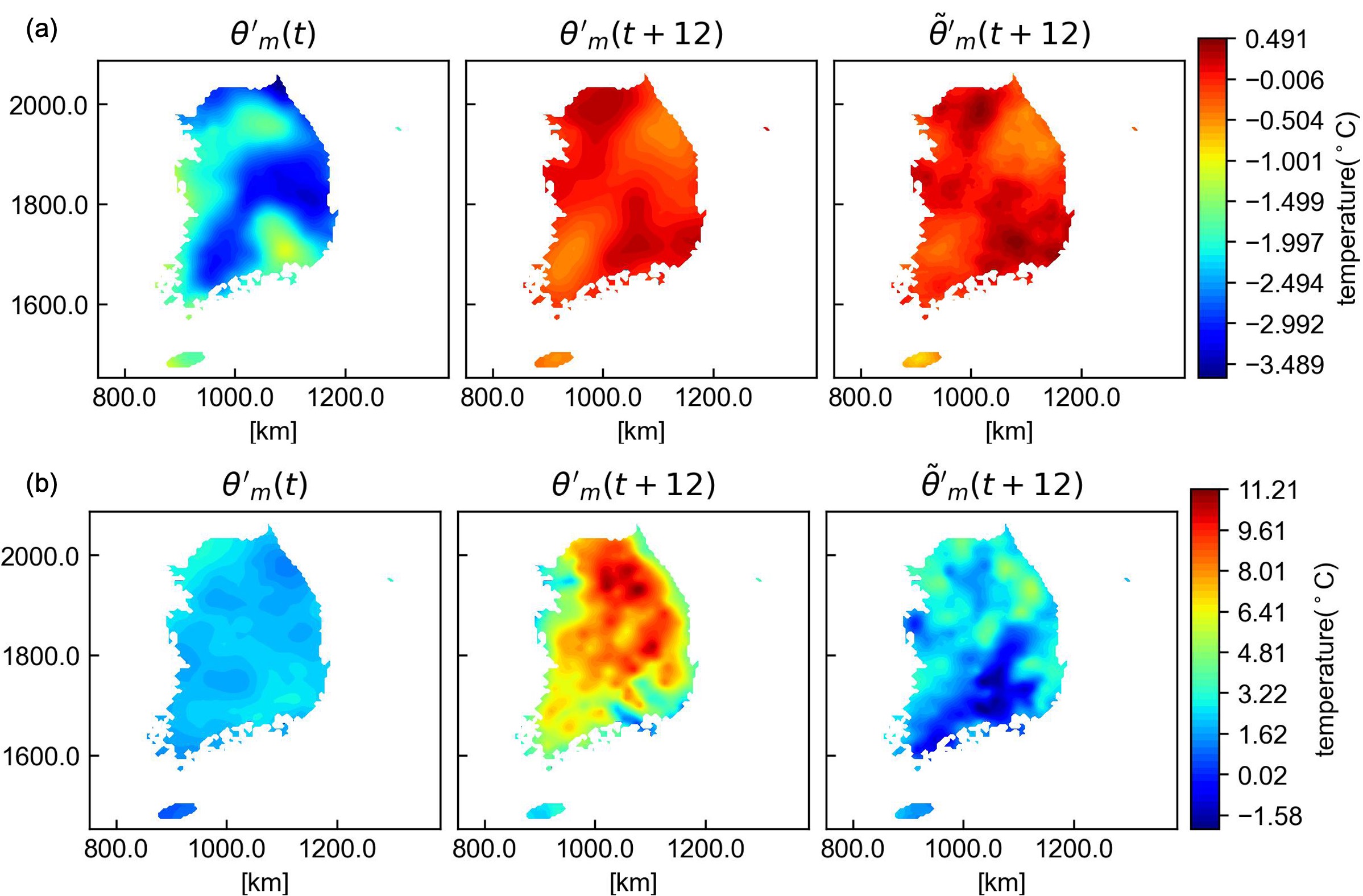}
\caption{Two sample test results by $TPTNet_{CNN}$ for 12-hour forecast. (a) the best prediction case at 2009/10/10 1:00 pm and (b) the worst prediction case at 2005/1/7 5:00 am among all the test cases.}
\label{fig:contours_12h}
\end{figure}

Among the test cases, examples of the best and worst predictions for 12-hour forecast by $TPTNet_{CNN}$ are shown in Figures \ref{fig:contours_12h}(a) and \ref{fig:contours_12h}(b), respectively. The first column shows the input data, the second column shows the target data after 12 h, and the third column shows the network prediction results. In the best prediction, the detailed distribution of temperature was well captured, whereas in the worst prediction, the overall temperature increase by 10 $^\circ C$ over 12 h was not well predicted. Given that this sudden increase occurs in $\theta'_m$ after eliminating the daily variation, such a sudden increase in temperature during the winter season is difficult to predict.

\subsection{Test of TPTNet for the unseen data of 2020}
\begin{figure}
\centering
\noindent\includegraphics[width=0.6\textwidth]{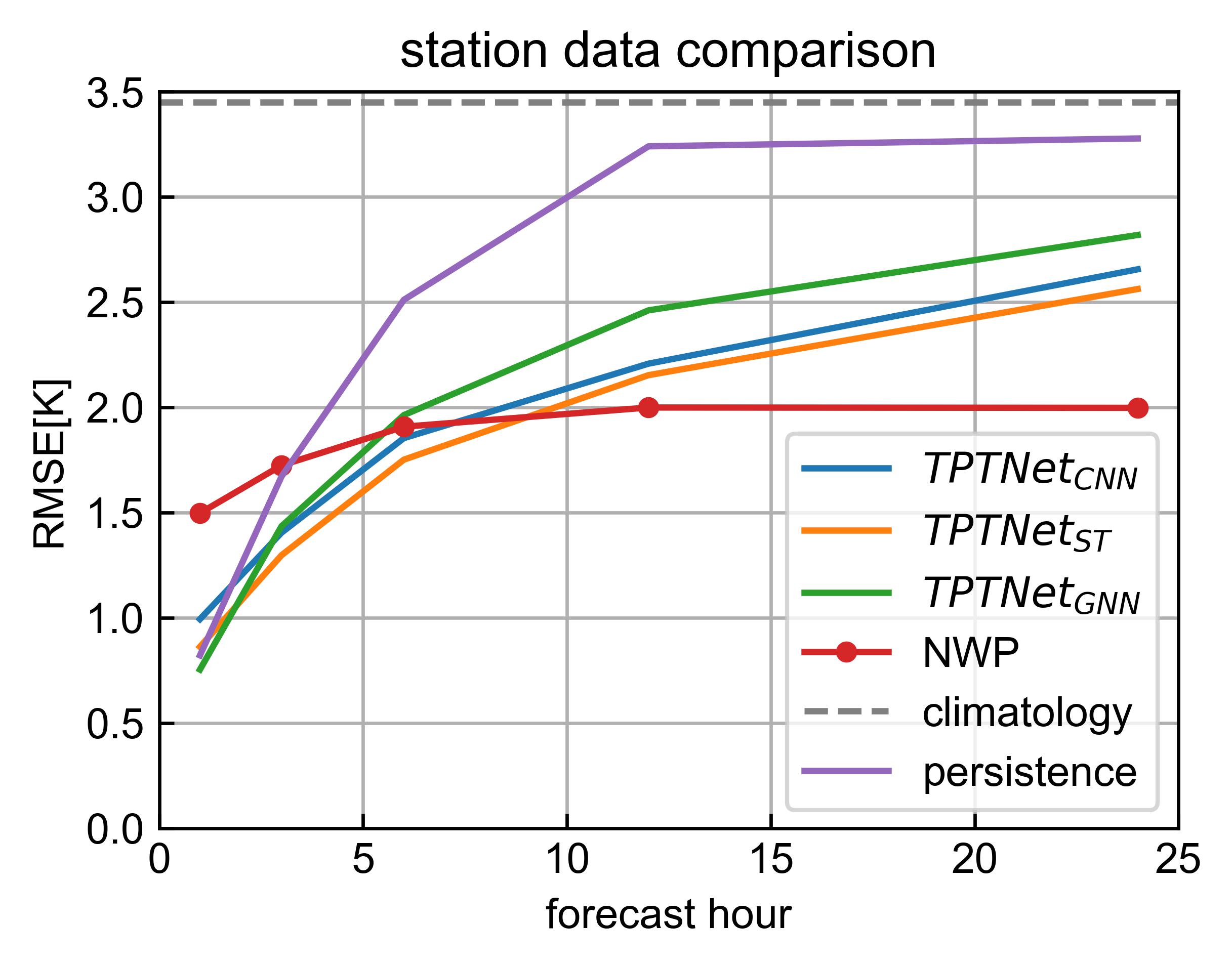}
\caption{Test results for the unseen data of 2020 for 5 forecast hours: 1, 3, 6, 12 and 24 hours. RMSEs by $TPTNet_{CNN}$, $TPTNet_{ST}$, $TPTNet_{GNN}$, NWP which is LDAPS of Korea, persistence, and climatology are plotted.}
\label{fig:station_comparison_2020_hs}
\end{figure}

In this section, the performances of the three trained networks based on data from 2000 to 2019 are evaluated for unseen data from 2020. To ensure consistency with the real circumstances of weather forecasts, the mean temperature $\langle{T\rangle}(t)$ obtained using data up to 2019 was used in the decomposition of the temperature data for 2020. RMSE of the temperature at the stations by the three network models was compared against the NWP data and the persistence and climatology forecasts for five forecast hours in Figure \ref{fig:station_comparison_2020_hs}.  For lead times of up to 6 h, all three models outperform NWP, while for 24-hour prediction, RMSE for all three models is larger than that of NWP by more than 0.5K. For 12-hour prediction, only $TPTNet_{CNN}$ and $TPTNet_{ST}$ perform comparably to NWP. $TPTNet_{GNN}$, which is free of interpolation errors, outperforms the other models only for the 1-hour prediction. We also tried rollout predictions, such as two consecutive applications of the 6-hour prediction model or four applications of the 3-hour prediction model for the 12-hour prediction, but they did not produce a better prediction than a single prediction. This was expected because the error curve of a single prediction is a concave function of the forecast hour, as shown in Figure \ref{fig:station_comparison_2020_hs}. The accumulated errors of the rollout predictions were likely to be larger than those of a single prediction. Given that the half-averaged integral time scale was 17 h, our models predicted the temperature well up to 12 h on average. 

\begin{figure}
\noindent\includegraphics[width=\textwidth]{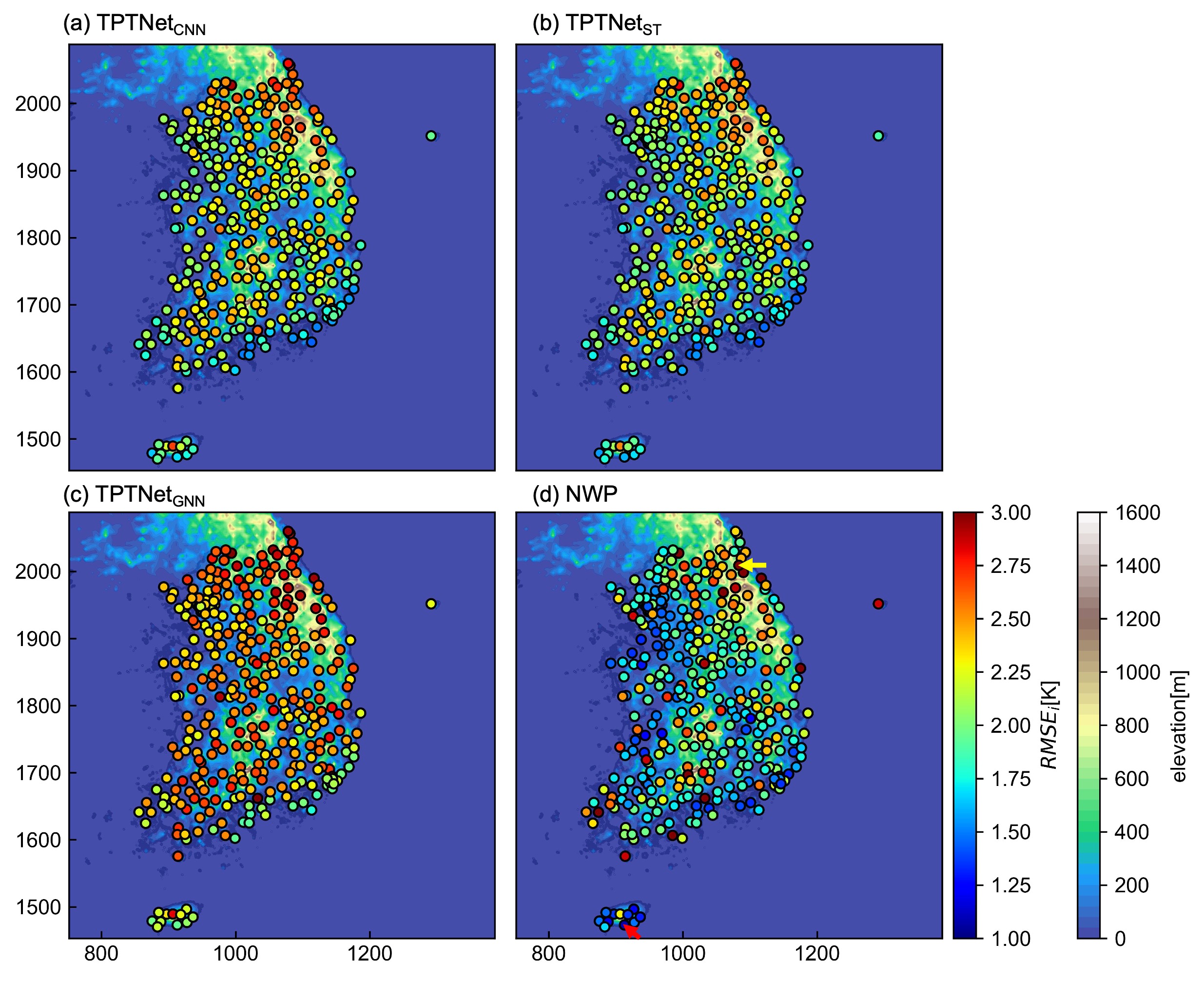}
\caption{Comparison of 12-hour predictions at stations by (a) $TPTNet_{CNN}$, (b) $TPTNet_{ST}$, (c) $TPTNet_{GNN}$ and (d) NWP. The average of $RMSE_i$ of (a), (b), (c) and (d) are 2.21 K, 2.15 K, 2.46 K and 2.00 K, respectively. Those with RMSE greater than 3K are marked in red. The yellow and red arrows point to Yangyang station (ID 596) with the largest RMSE and Seoguipo station (ID 189) with the smallest RMSE by NWP, respectively.}
\label{fig:h12}
\end{figure}

Because RMSE, which is the station-averaged error, alone does not provide sufficient information on the performance, we investigated the detailed distribution of $RMSE_i$ over all stations, focusing on 12-hour prediction case, as shown in Figure \ref{fig:h12}. $RMSE_i$ values of all three models were compared with those of NWP. The distributions of $RMSE_i$ by $TPTNet_{CNN}$ and $TPTNet_{ST}$ were similar, whereas $RMSE_i$ by $TPTNet_{GNN}$ was relatively uniform across the stations. Comparing the three models, $RMSE_i$ by NWP was widely scattered over the stations, although the mean values were comparable. Generally, the errors at stations located in the northern part near the East coast were large, whereas those near the West and South coasts were relatively small. One interesting observation is that the error at the station in Ulleung Island in East Sea, where only one station is operated, by our models is small at approximately 2K, whereas that by NWP is large at 3K.

\begin{figure}
\noindent\includegraphics[width=\textwidth]{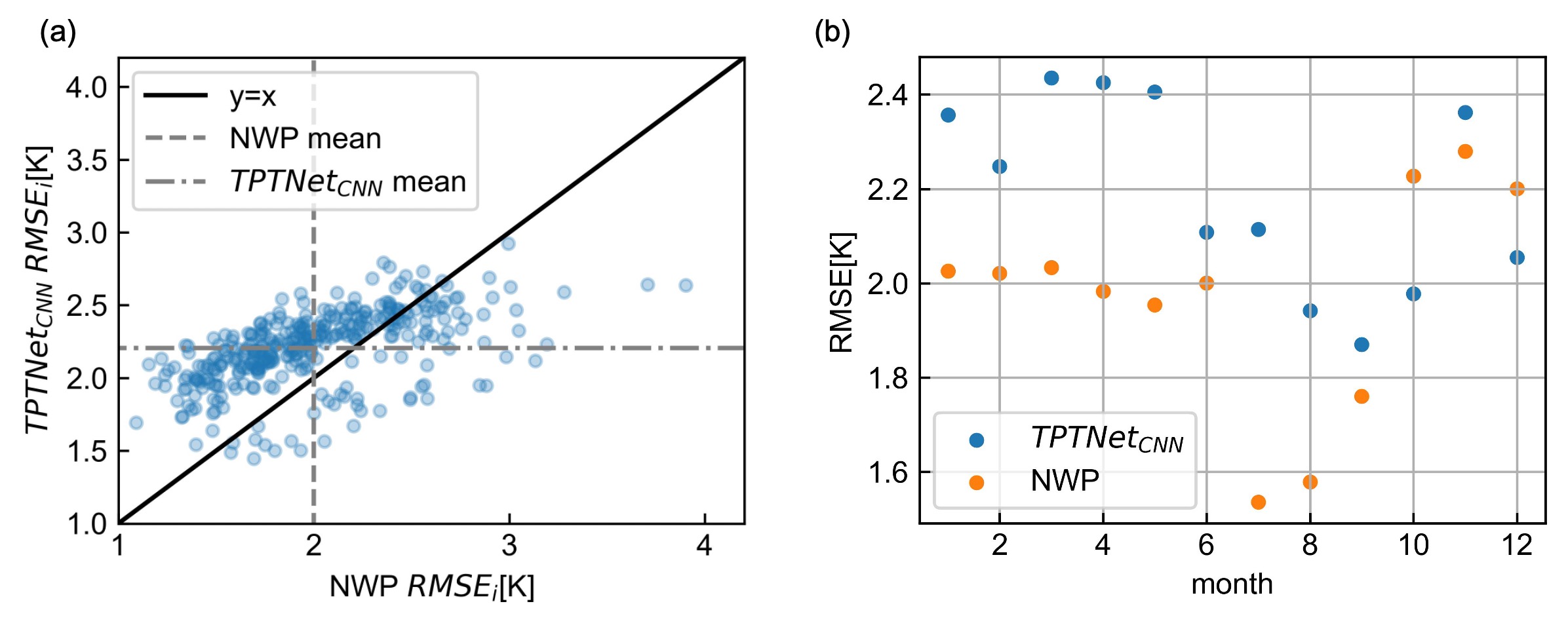}
\caption{Comparison of 12-hour predictions by $TPTNet_{CNN}$ and NWP. (a) Scatter plot between $RMSE_i$ at each station between $TPTNet_{CNN}$ and NWP, and (b) the monthly averaged $RMSE$ of all stations. In (a), the mean values of $RMSE$ by the two models are shown in a gray line. The average $RMSE$ of $TPTNet_{CNN}$ and NWP  is 2.21 K and 2.0 K, respectively.}
\label{fig:h12_1}
\end{figure}

The scatter plot of $RMSE_i$ between NWP and $TPTNet_{CNN}$ shown in Figure. \ref{fig:h12_1}(a), clearly indicates that the errors of NWP are more scattered than those of $TPTNet_{CNN}$ although the mean errors are similar. The range of $RMSE_i$ of NWP is between 1K and 4K, whereas $RMSE_i$ of $TPTNet_{CNN}$ ranges from 1.5K to 3K. From this observation, we can say that the quality of the prediction by $TPTNet_{CNN}$ is better and thus more reliable than NWP because $TPTNet_{CNN}$ produces the worst predictions less frequently. Figure \ref{fig:h12_1}(b) compares the monthly average RMSE of NWP and $TPTNet_{CNN}$. For both models, the predictions for summer were better than those for the other seasons. For September and December, $TPTNet_{CNN}$ outperformed NWP. A similar trend was observed for the 1, 3, 6, and 24-hour forecasts, which are provided in the Supporting Information (Figure S3 to S10).

\begin{figure}
\noindent\includegraphics[width=\textwidth]{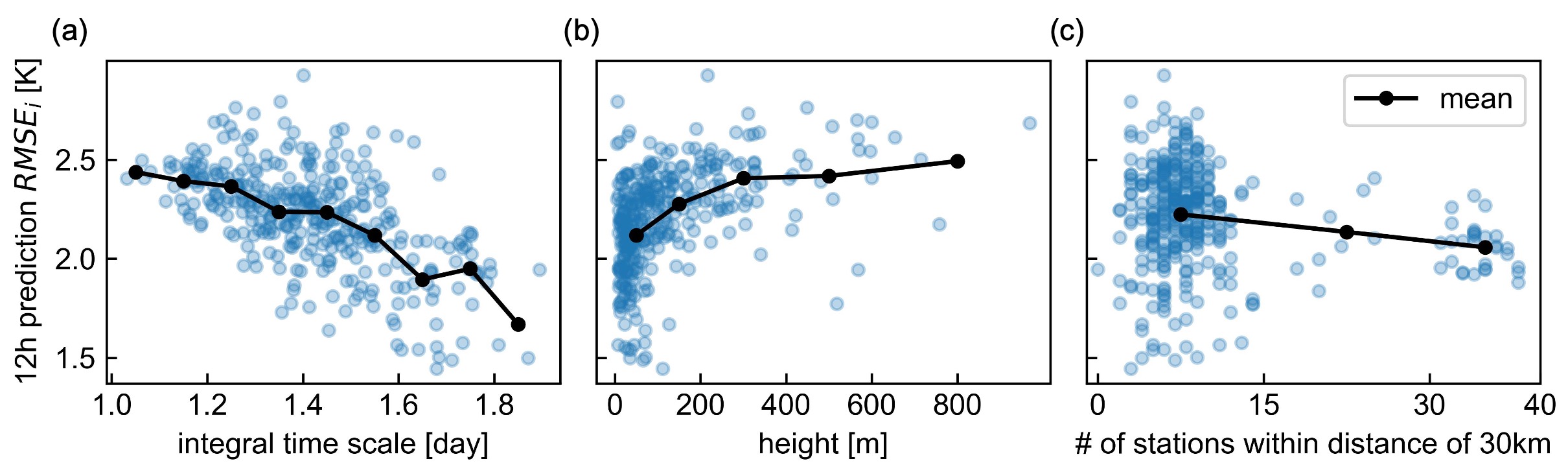}
\caption{Scatter plots of $RMSE_i$ by $TPTNet_{CNN}$ for 12-hour prediction versus (a) the integral time scale at each station, (b) the height of stations, and (c) the number of stations within 30km. Black solid line with symbol denotes the interval average value.}
\label{fig:h12_2}
\end{figure}

For a more detailed investigation of the prediction errors, scatter plots of $RMSE_i$ by $TPTNet_{CNN}$ against the integral time scale of the temperature, height of stations, and number of stations within 30km are illustrated in Figure \ref{fig:h12_2}. In our construction of prediction networks, we hypothesized that short-term local forecasts are closely dependent on the integral timescale. Figure \ref{fig:h12_2}(a) confirms this hypothesis because it shows that the larger the ITS at the station, the smaller the $RMSE_i$. The correlation between the altitude of the station and $RMSE_i$ is shown in Figure \ref{fig:h12_2}(b), which indicates that the prediction at a higher station is more difficult, although stations located above 500m are rare. In another trial, we attempted to incorporate height information into the network training; however, no significant improvement was achieved. The correlation between station density and $RMSE_i$ is shown in Figure \ref{fig:h12_2}(c). Station density was quantified as the number of stations within 30km. For most stations, the nearby station density was below 15, with two groups of exceptions: one group with $15 \sim 30$ and the other group with $30 \sim 40$. The most populated group corresponded to Seoul, followed by Seoul neighborhoods. The mean error for Seoul was smaller than that for other regions, suggesting that the installation of more stations might lead to better predictions.

\begin{figure}
\noindent\includegraphics[width=\textwidth]{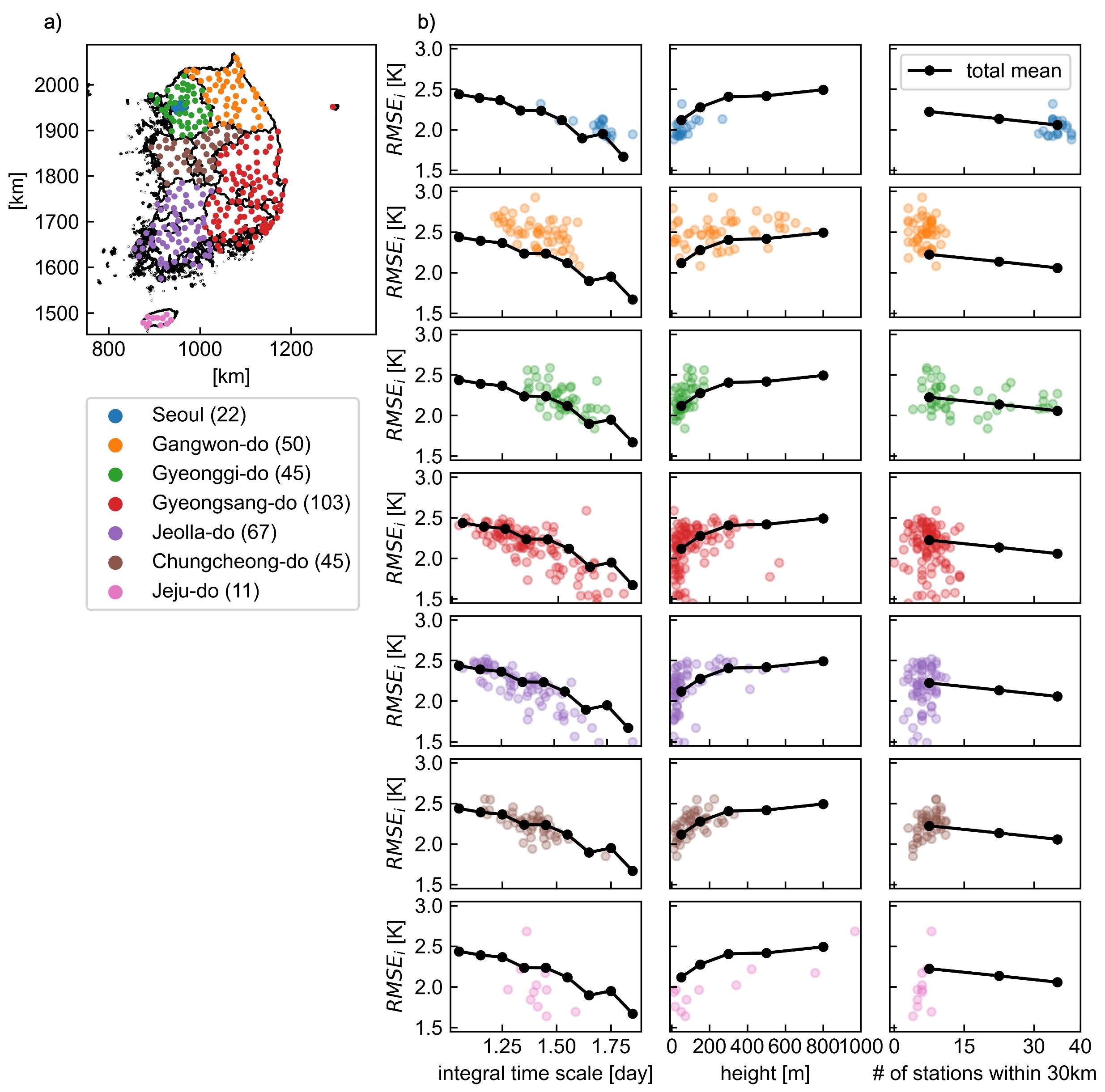}
\caption{Regional contribution to the scatter plot of $RMSE_i$ by $TPTNet_{CNN}$ for 12-hour prediction shown in Figure \ref{fig:h12_2}. (a) Stations in each province are marked in different colors and (b) station data in each province are shown in the corresponding color. The number in the legend denotes the number of stations in each province. Black solid line with symbol denotes the average value of total data.}
\label{fig:h12_3}
\end{figure}

The regional contributions to the scatterplots shown in Figure \ref{fig:h12_2} are investigated in Figure \ref{fig:h12_3}, which would be useful for policy decisions to improve forecasts. The low prediction error in the Seoul area was mostly due to the high density of stations (22 stations within Seoul). In Gyeonggi-do region surrounding Seoul, the stations are relatively densely distributed, leading to a low prediction error. However, the prediction error of Gangwon-do region was the highest among all provinces, primarily because most stations were located in mountains, as shown in the height distribution. Even at stations near the coast, as identified by the height in this region, the errors were significant. In most other provinces, the trend of the scatter plot is similar to that of the total data shown in Figure \ref{fig:h12_2}, whereas Jeju-do region, which is an island, exhibits an exceptional distribution. Although the stations were located at high altitudes and were relatively sparsely distributed, the errors were smaller than average, except for one station located at an altitude of almost 1,000m. It is conjectured that the simple cone-shaped topography of Jeju Island with a mountain at the center might be responsible for the more predictable behavior, and that the weather is less influenced by the complicated pattern of weather on the mainland.

\begin{figure}
\noindent\includegraphics[width=\textwidth]{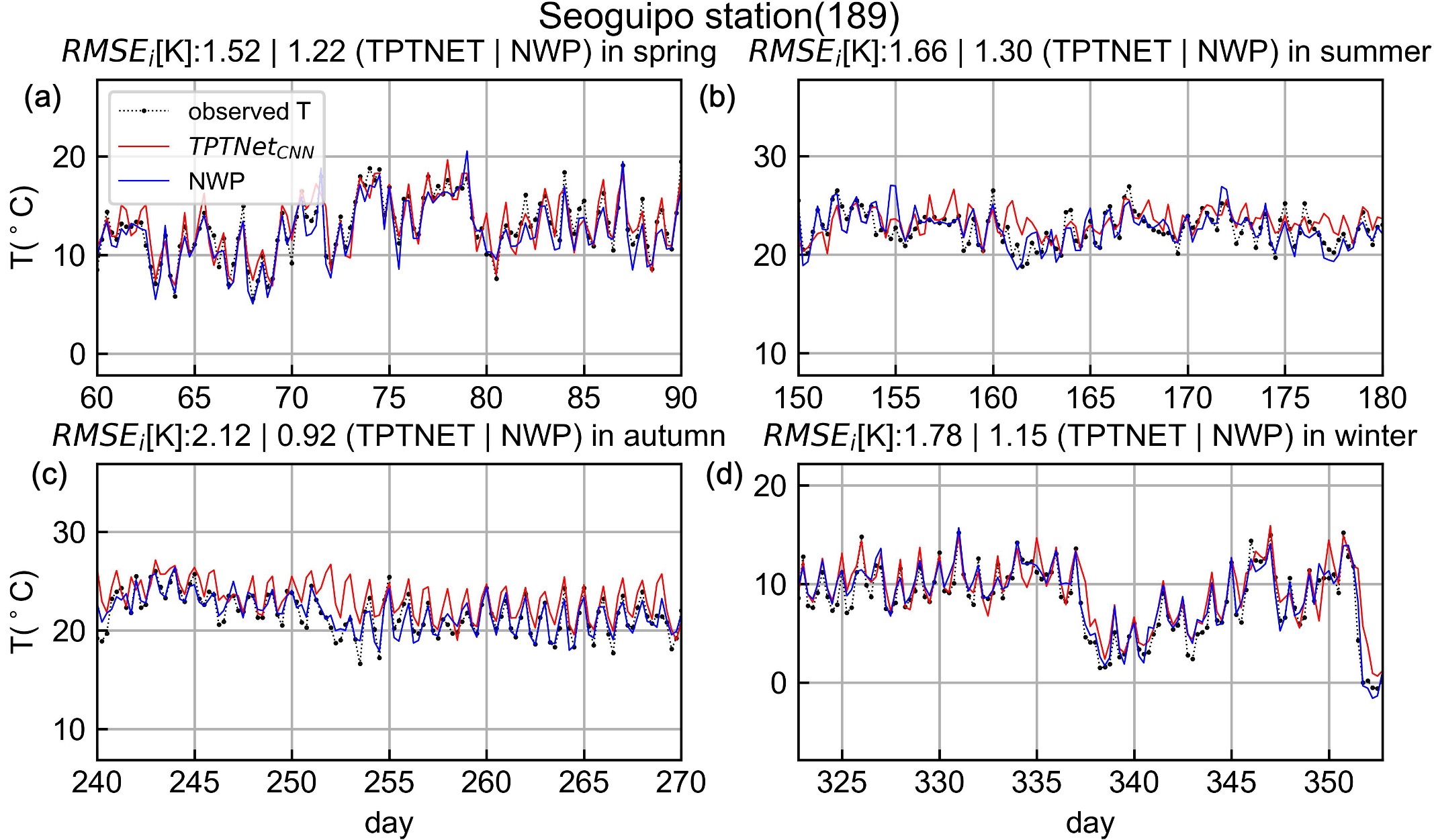}
\caption{Comparison of the predictions by $TPTNet_{CNN}$ and NWP for 12-hour prediction against observation at Seoguipo station, where NWP prediction is the best, indicated by the red arrow in Figure \ref{fig:h12}(b). Time series at 6-hour resolution is compared for one-month period in (a) spring, (b) summer, (c) autumn and (d) winter.}
\label{fig:s189}
\end{figure}

\begin{figure}
\noindent\includegraphics[width=\textwidth]{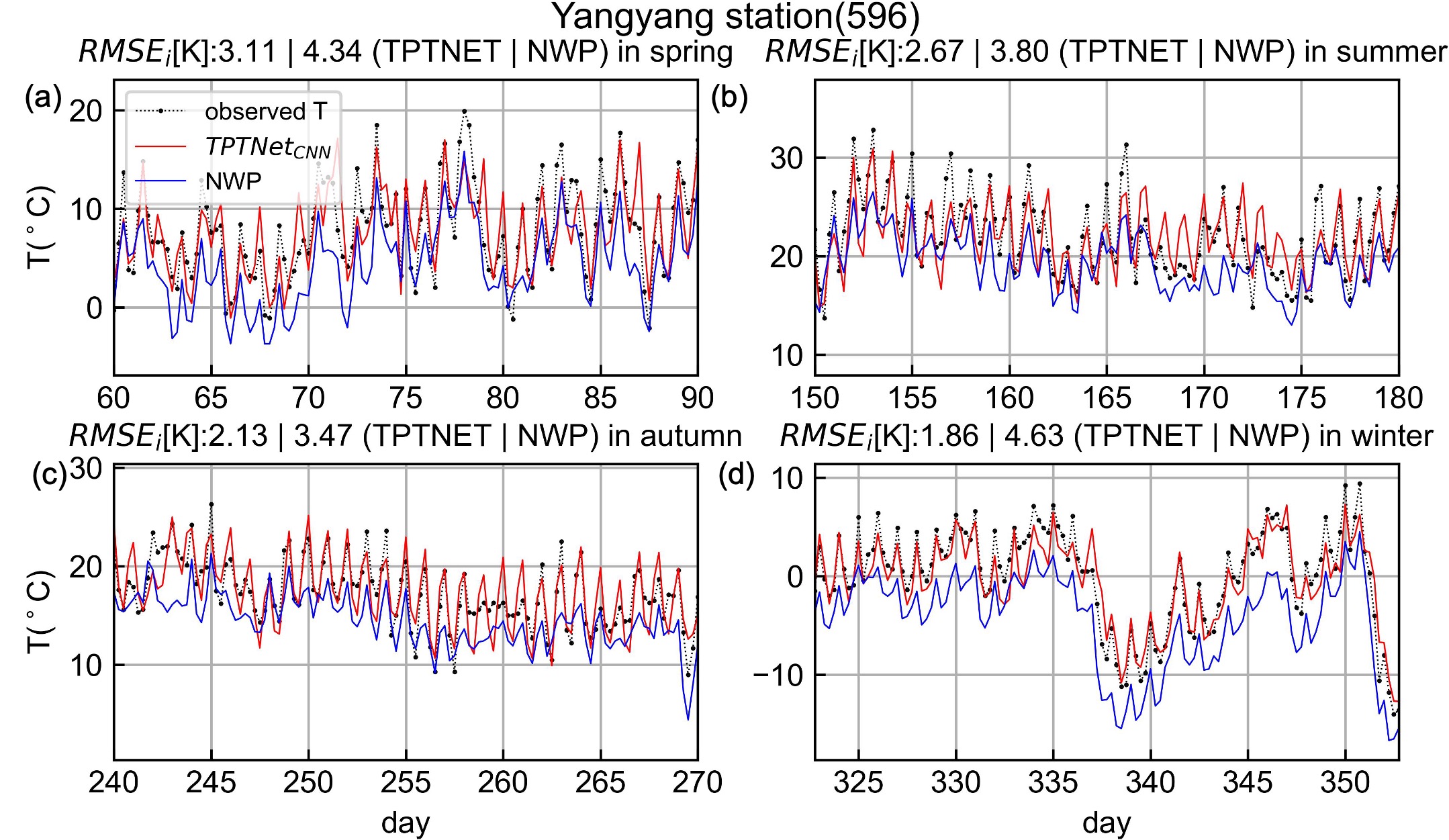}
\caption{Comparison of the predictions by $TPTNet_{CNN}$ and NWP for 12-hour prediction against observation at Yangyang station, where NWP prediction is the worst, indicated by the yellow arrow in Figure \ref{fig:h12}(b). Time series at 6-hour resolution is compared for one-month period in (a) spring, (b) summer, (c) autumn and (d) winter.}
\label{fig:s596}
\end{figure}

Figures \ref{fig:s189} and \ref{fig:s596} illustrate comparison examples of time-series prediction by NWP and $TPTNet_{CNN}$ against station measurements at two stations (ID 189 and 596), respectively. These two stations were selected because stations 189 and 596 yielded the best and worst NWP predictions, respectively. The locations are marked by yellow and red arrows in Figure \ref{fig:h12}(d). These plots display the observed $T$ along with the prediction results using $TPTNet_{CNN}$ and NWP at a 6-hour resolution for each season, covering approximately one month of data. The RMSE between the observed $T$ and prediction result in the corresponding interval was calculated and displayed in the title of each plot. For station 189, the predictions of $TPTNet_{CNN}$ and NWP did not exhibit significant deviations from the observed data, whereas noticeable discrepancies were observed between $TPTNet_{CNN}$ and NWP for station 596, and $TPTNet_{CNN}$ outperforms NWP significantly. Given that station 596 is located in a mountainous region, the wide variation in temperature during the spring season was captured well by $TPTNet_{CNN}$. Although this is only one example, an investigation of other stations exhibiting large NWP errors confirms that this is the case.

\begin{figure}
    \noindent\includegraphics[width=\textwidth]{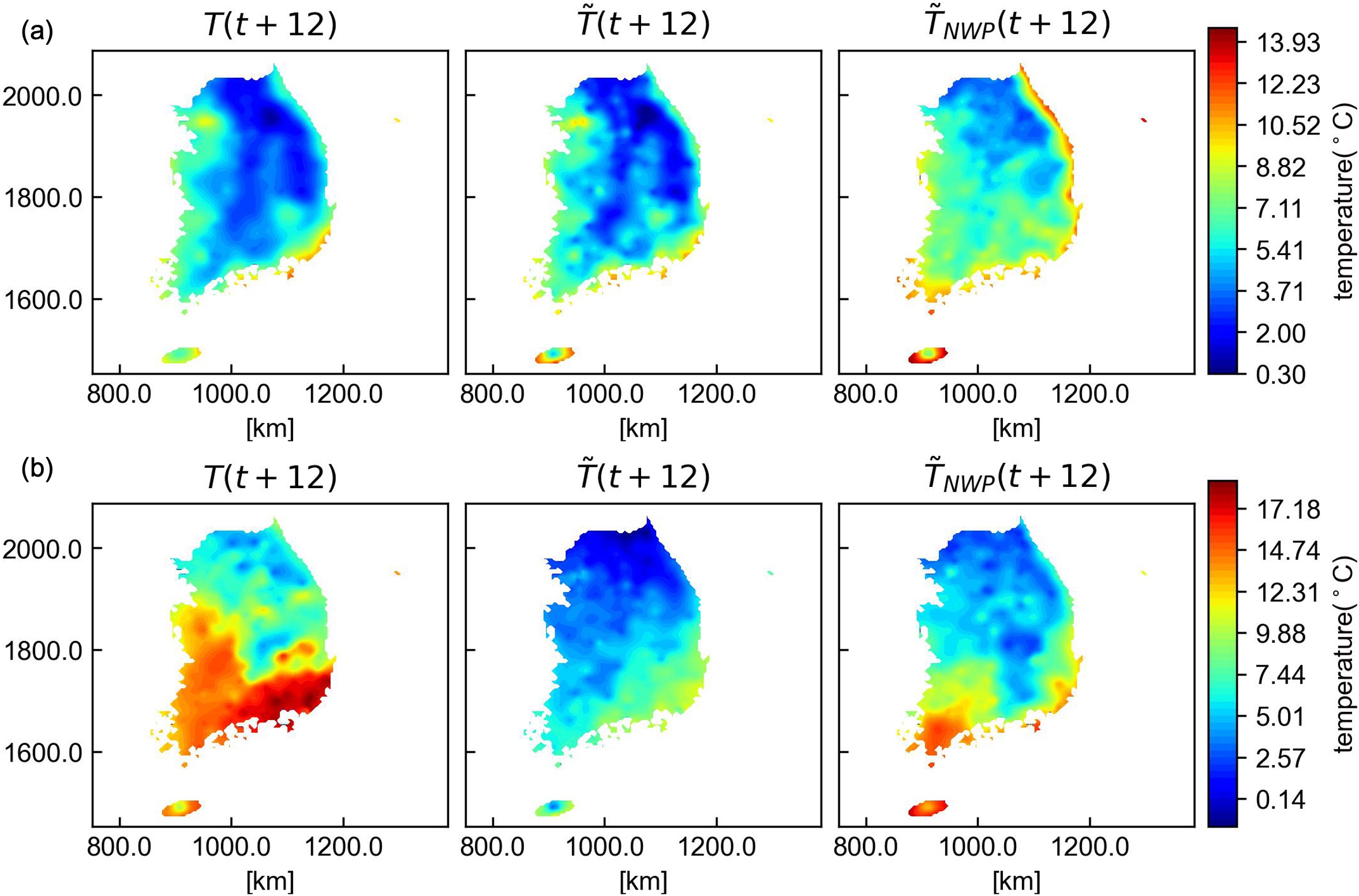}
    \caption{Two example test results of 12-hour forecast at (a) 2020/11/1 9:00 pm (the best prediction by $TPTNet_{CNN}$) and (b) 2020/1/7 9:00 pm (the worst prediction by $TPTNet_{CNN}$).}
    \label{fig:tptNet_vs_nwp}
\end{figure}

Finally, Figure \ref{fig:tptNet_vs_nwp} compares the predicted temperature distribution over the entire region by 12-hour prediction by $TPTNet_{CNN}$ and NWP against the observation data at two instances. In the first instance shown in Figure \ref{fig:tptNet_vs_nwp}(a), the prediction by $TPTNet_{CNN}$ is the best, and the second instance in Figure \ref{fig:tptNet_vs_nwp} corresponds to the worst case. For a fair comparison, ordinary kriging was performed on the observation data at the stations and NWP predictions at different meshes to generate mesh data. The prediction by $TPTNet_{CNN}$ in the first instance was excellent for capturing detailed small-scale variations, whereas the prediction by NWP was not as accurate as $TPTNet_{CNN}$. In the second instance, the predictions by $TPTNet_{CNN}$ and NWP are poor. Wild variation in temperature over the South Korean Peninsula by more than 15 $^\circ C$ during the winter season (Jan. 7th 9am) is an extreme event that makes accurate prediction very difficult. The movies of T(t+12) and $\tilde{T}(t+12)$ for each month are available in the Supporting Information (Movies S1 to S12).

\section{Conclusions}

In this study, we developed a new deep-learning-based model TPTNet that can predict the local distribution of 2m temperature using only ground station data. Our models were trained to predict the temperature fluctuation component from the average yearly and daily variations called climatology extracted from 20-year station data of South Korea. The altitude of the ground stations was reflected by the introduction of the potential temperature. Three types of learning networks based on CNN, Swin Transformer, and GNN were trained using station data from 2000 to 2019 and tested for the unseen data of the year 2020. Our networks based on CNN and Swin Transformer outperformed NWP in the prediction of 2m temperature at stations for forecast hours of up to 12 h. GNN-based network slightly underperformed although the model was free of interpolation errors.

From the detailed investigation of the prediction error distribution over the stations for 12-hour forecast, we observed that while the errors by NWP exhibit wide variation over the stations ranging from 1K to 4K, our networks based on CNN and Swin Transformer produce relatively uniform errors between 1.5K and 3K, although the average errors around 2K are comparable, implying that our networks yield less poor predictions. We also observed that the error at stations with longer integral time scales was smaller and that the higher the altitude of a station, the larger the error at the station. In addition, the errors at stations with more nearby stations were smaller.

Certain meteorological variables contribute significantly to enhancing the prediction performance, whereas others do not significantly affect the prediction outcomes \cite{Gong2022}. In our study, we focused on prediction of 2m temperature based solely on 2m temperature. We attempted to test the effect of terrain height and relative humidity on the prediction performance by concatenating the information to the input, but found that neither variable significantly improved the performance. The effects of other meteorological variables such as pressure, wind speed, and wind direction should be investigated more extensively in the future.

Our networks were trained using station data only on the Korean Peninsula; thus, reasonable prediction was possible for short forecast hours of up to 12 h. For longer-term prediction of up to 48 h, station data over a wider region, including North Korea, Japan, and northeastern China, could be utilized, which is similar to the region covered by NWP model (LDAPS) considered in this study. Extension of the model domain may lead to not only reliable long-term prediction but also better short-term prediction at the expense of learning costs.

We demonstrated that the reliable prediction of 2m temperature for forecast hours up to 12 h based on station data only is possible through neural-network-based learning. Although training takes 30 h on a single GPU machine when using 20 years of data, prediction takes less than one minute, allowing for real-time prediction based on station data only. Furthermore, the quality of the prediction is better than NWP in that our networks produce fewer poor predictions.


%
%
%
%

%
%

%

%

\section*{Data Availability Statement}
The data used in this study are available at \citeA{DVN/J3DRDT_2023}. The code for this study is available at \citeA{park_2023_10252672}.

Temporary private links for the data and software are \url{https://dataverse.harvard.edu/privateurl.xhtml?token=52fd4330-1ec2-4808-9233-14aaa4aa00d7} and \url{https://zenodo.org/records/10252672?token=eyJhbGciOiJIUzUxMiIsImlhdCI6MTcwMjM3MDQ5OSwiZXhwIjoxNzM1NjAzMTk5fQ.eyJpZCI6ImU5OWY2MjcyLWQ2ZDgtNDZmZC1hOGVmLTZiYjg4ODZkN2JiMCIsImRhdGEiOnt9LCJyYW5kb20iOiI0NWYzZDI5ZGQ4ZmQ2NWM0OTFjNWE1N2Y4ZDhiNTZkNSJ9.gYdr3LI7DnDj5_bvR15njUojjNX4jmuyOiVH8UMv8zUcJynTfU5dZU5gEv8TlbwfjI01ccANmRJYKdNPmHVzSg}, respectively.







\acknowledgments
This work was supported by the National Research Foundation of Korea (NRF) grant funded by the Korean Government (MSIP) (2022R1A2C2005538).



%
%



\bibliography{library}

\begin{thebibliography}{}

\bibitem [\protect \citeauthoryear {%
Azari%
, Hassan%
, Pierce%
\BCBL {}\ \BBA {} Ebrahimi%
}{%
Azari%
\ \protect \BOthers {.}}{%
{\protect \APACyear {2022}}%
}]{%
Azari2022}
\APACinsertmetastar {%
Azari2022}%
\begin{APACrefauthors}%
Azari, B.%
, Hassan, K.%
, Pierce, J.%
\BCBL {}\ \BBA {} Ebrahimi, S.%
\end{APACrefauthors}%
\unskip\
\newblock
\APACrefYearMonthDay{2022}{}{}.
\newblock
{\BBOQ}\APACrefatitle {Evaluation of Machine Learning Methods Application in
  Temperature Prediction} {Evaluation of machine learning methods application
  in temperature prediction}.{\BBCQ}
\newblock
\APACjournalVolNumPages{Computational Research Progress in Applied Science \&
  Engineering}{8}{}{1--12}.
\newblock
\begin{APACrefDOI} \doi{10.52547/crpase.8.1.2747} \end{APACrefDOI}
\PrintBackRefs{\CurrentBib}

\bibitem [\protect \citeauthoryear {%
Belkin%
, Hsu%
, Ma%
\BCBL {}\ \BBA {} Mandal%
}{%
Belkin%
\ \protect \BOthers {.}}{%
{\protect \APACyear {2019}}%
}]{%
Belkin2019}
\APACinsertmetastar {%
Belkin2019}%
\begin{APACrefauthors}%
Belkin, M.%
, Hsu, D.%
, Ma, S.%
\BCBL {}\ \BBA {} Mandal, S.%
\end{APACrefauthors}%
\unskip\
\newblock
\APACrefYearMonthDay{2019}{}{}.
\newblock
{\BBOQ}\APACrefatitle {Reconciling modern machine-learning practice and the
  classical bias--variance trade-off} {Reconciling modern machine-learning
  practice and the classical bias--variance trade-off}.{\BBCQ}
\newblock
\APACjournalVolNumPages{Proceedings of the National Academy of Sciences of the
  United States of America}{116}{}{15849--15854}.
\newblock
\begin{APACrefDOI} \doi{10.1073/pnas.1903070116} \end{APACrefDOI}
\PrintBackRefs{\CurrentBib}

\bibitem [\protect \citeauthoryear {%
Gong%
\ \protect \BOthers {.}}{%
Gong%
\ \protect \BOthers {.}}{%
{\protect \APACyear {2022}}%
}]{%
Gong2022}
\APACinsertmetastar {%
Gong2022}%
\begin{APACrefauthors}%
Gong, B.%
, Langguth, M.%
, Ji, Y.%
, Mozaffari, A.%
, Stadtler, S.%
, Mache, K.%
\BCBL {}\ \BBA {} Schultz, M\BPBI G.%
\end{APACrefauthors}%
\unskip\
\newblock
\APACrefYearMonthDay{2022}{}{}.
\newblock
{\BBOQ}\APACrefatitle {Temperature forecasting by deep learning methods}
  {Temperature forecasting by deep learning methods}.{\BBCQ}
\newblock
\APACjournalVolNumPages{Geoscientific Model Development}{15}{23}{8931--8956}.
\PrintBackRefs{\CurrentBib}

\bibitem [\protect \citeauthoryear {%
Hersbach%
\ \protect \BOthers {.}}{%
Hersbach%
\ \protect \BOthers {.}}{%
{\protect \APACyear {2020}}%
}]{%
Hersbach2020}
\APACinsertmetastar {%
Hersbach2020}%
\begin{APACrefauthors}%
Hersbach, H.%
, Bell, B.%
, Berrisford, P.%
, Hirahara, S.%
, Hor{\'a}nyi, A.%
, Mu{\~n}oz-Sabater, J.%
\BDBL {}Th{\'e}paut, J\BPBI N.%
\end{APACrefauthors}%
\unskip\
\newblock
\APACrefYearMonthDay{2020}{}{}.
\newblock
{\BBOQ}\APACrefatitle {The ERA5 global reanalysis} {The era5 global
  reanalysis}.{\BBCQ}
\newblock
\APACjournalVolNumPages{Quarterly Journal of the Royal Meteorological
  Society}{146}{}{1999--2049}.
\newblock
\begin{APACrefDOI} \doi{10.1002/qj.3803} \end{APACrefDOI}
\PrintBackRefs{\CurrentBib}

\bibitem [\protect \citeauthoryear {%
Hou%
, Wang%
, Zhou%
\BCBL {}\ \BBA {} Tian%
}{%
Hou%
\ \protect \BOthers {.}}{%
{\protect \APACyear {2022}}%
}]{%
Hou2022}
\APACinsertmetastar {%
Hou2022}%
\begin{APACrefauthors}%
Hou, J.%
, Wang, Y.%
, Zhou, J.%
\BCBL {}\ \BBA {} Tian, Q.%
\end{APACrefauthors}%
\unskip\
\newblock
\APACrefYearMonthDay{2022}{}{}.
\newblock
{\BBOQ}\APACrefatitle {Prediction of hourly air temperature based on CNN--LSTM}
  {Prediction of hourly air temperature based on cnn--lstm}.{\BBCQ}
\newblock
\APACjournalVolNumPages{Geomatics, Natural Hazards and Risk}{13}{}{1962--1986}.
\newblock
\begin{APACrefDOI} \doi{10.1080/19475705.2022.2102942} \end{APACrefDOI}
\PrintBackRefs{\CurrentBib}

\bibitem [\protect \citeauthoryear {%
Hu%
, Chen%
, Wang%
\BCBL {}\ \BBA {} Li%
}{%
Hu%
\ \protect \BOthers {.}}{%
{\protect \APACyear {2023}}%
}]{%
Hu2023}
\APACinsertmetastar {%
Hu2023}%
\begin{APACrefauthors}%
Hu, Y.%
, Chen, L.%
, Wang, Z.%
\BCBL {}\ \BBA {} Li, H.%
\end{APACrefauthors}%
\unskip\
\newblock
\APACrefYearMonthDay{2023}{}{}.
\newblock
{\BBOQ}\APACrefatitle {SwinVRNN: A Data-Driven Ensemble Forecasting Model via
  Learned Distribution Perturbation} {Swinvrnn: A data-driven ensemble
  forecasting model via learned distribution perturbation}.{\BBCQ}
\newblock
\APACjournalVolNumPages{Journal of Advances in Modeling Earth Systems}{}{}{}.
\newblock
\begin{APACrefURL} \url{http://arxiv.org/abs/2205.13158} \end{APACrefURL}
\newblock
\begin{APACrefDOI} \doi{10.1029/2022MS003211} \end{APACrefDOI}
\PrintBackRefs{\CurrentBib}

\bibitem [\protect \citeauthoryear {%
Isaaks%
\ \BBA {} Srivastava%
}{%
Isaaks%
\ \BBA {} Srivastava%
}{%
{\protect \APACyear {1989}}%
}]{%
Isaaks1989}
\APACinsertmetastar {%
Isaaks1989}%
\begin{APACrefauthors}%
Isaaks, E\BPBI H.%
\BCBT {}\ \BBA {} Srivastava, R\BPBI M.%
\end{APACrefauthors}%
\unskip\
\newblock
\APACrefYear{1989}.
\newblock
\APACrefbtitle {Applied geostatistics} {Applied geostatistics}.
\newblock
\APACaddressPublisher{}{Oxford University Press}.
\PrintBackRefs{\CurrentBib}

\bibitem [\protect \citeauthoryear {%
Karras%
, Aila%
, Laine%
\BCBL {}\ \BBA {} Lehtinen%
}{%
Karras%
\ \protect \BOthers {.}}{%
{\protect \APACyear {2017}}%
}]{%
karras2017}
\APACinsertmetastar {%
karras2017}%
\begin{APACrefauthors}%
Karras, T.%
, Aila, T.%
, Laine, S.%
\BCBL {}\ \BBA {} Lehtinen, J.%
\end{APACrefauthors}%
\unskip\
\newblock
\APACrefYearMonthDay{2017}{}{}.
\newblock
{\BBOQ}\APACrefatitle {Progressive growing of gans for improved quality,
  stability, and variation} {Progressive growing of gans for improved quality,
  stability, and variation}.{\BBCQ}
\newblock
\APACjournalVolNumPages{arXiv preprint arXiv:1710.10196}{}{}{}.
\PrintBackRefs{\CurrentBib}

\bibitem [\protect \citeauthoryear {%
Keisler%
}{%
Keisler%
}{%
{\protect \APACyear {2022}}%
}]{%
Keisler2022}
\APACinsertmetastar {%
Keisler2022}%
\begin{APACrefauthors}%
Keisler, R.%
\end{APACrefauthors}%
\unskip\
\newblock
\APACrefYearMonthDay{2022}{}{}.
\newblock
{\BBOQ}\APACrefatitle {Forecasting global weather with graph neural networks}
  {Forecasting global weather with graph neural networks}.{\BBCQ}
\newblock
\APACjournalVolNumPages{arXiv preprint arXIv:2202.07575v1}{}{}{}.
\PrintBackRefs{\CurrentBib}

\bibitem [\protect \citeauthoryear {%
H.~Kim%
, Kim%
\BCBL {}\ \BBA {} Lee%
}{%
H.~Kim%
\ \protect \BOthers {.}}{%
{\protect \APACyear {2023}}%
}]{%
Kim2023heat2}
\APACinsertmetastar {%
Kim2023heat2}%
\begin{APACrefauthors}%
Kim, H.%
, Kim, J.%
\BCBL {}\ \BBA {} Lee, C.%
\end{APACrefauthors}%
\unskip\
\newblock
\APACrefYearMonthDay{2023}{}{}.
\newblock
{\BBOQ}\APACrefatitle {{Interpretable deep learning for prediction of Prandtl
  number effect in turbulent heat transfer}} {{Interpretable deep learning for
  prediction of Prandtl number effect in turbulent heat transfer}}.{\BBCQ}
\newblock
\APACjournalVolNumPages{Journal of Fluid Mechanics}{955}{}{A14}.
\PrintBackRefs{\CurrentBib}

\bibitem [\protect \citeauthoryear {%
H.~Kim%
, Kim%
, Won%
\BCBL {}\ \BBA {} Lee%
}{%
H.~Kim%
\ \protect \BOthers {.}}{%
{\protect \APACyear {2021}}%
}]{%
Kim2021sup}
\APACinsertmetastar {%
Kim2021sup}%
\begin{APACrefauthors}%
Kim, H.%
, Kim, J.%
, Won, S.%
\BCBL {}\ \BBA {} Lee, C.%
\end{APACrefauthors}%
\unskip\
\newblock
\APACrefYearMonthDay{2021}{}{}.
\newblock
{\BBOQ}\APACrefatitle {Unsupervised deep learning for super-resolution
  reconstruction of turbulence} {Unsupervised deep learning for
  super-resolution reconstruction of turbulence}.{\BBCQ}
\newblock
\APACjournalVolNumPages{Journal of Fluid Mechanics}{}{}{}.
\newblock
\begin{APACrefDOI} \doi{10.1017/jfm.2020.1028} \end{APACrefDOI}
\PrintBackRefs{\CurrentBib}

\bibitem [\protect \citeauthoryear {%
J.~Kim%
, Kim%
, Kim%
\BCBL {}\ \BBA {} Lee%
}{%
J.~Kim%
\ \protect \BOthers {.}}{%
{\protect \APACyear {2022}}%
}]{%
Kim2022les}
\APACinsertmetastar {%
Kim2022les}%
\begin{APACrefauthors}%
Kim, J.%
, Kim, H.%
, Kim, J.%
\BCBL {}\ \BBA {} Lee, C.%
\end{APACrefauthors}%
\unskip\
\newblock
\APACrefYearMonthDay{2022}{}{}.
\newblock
{\BBOQ}\APACrefatitle {Deep reinforcement learning for large-eddy simulation
  modeling in wall-bounded turbulence} {Deep reinforcement learning for
  large-eddy simulation modeling in wall-bounded turbulence}.{\BBCQ}
\newblock
\APACjournalVolNumPages{Physics of Fluids}{34}{}{105132}.
\PrintBackRefs{\CurrentBib}

\bibitem [\protect \citeauthoryear {%
J.~Kim%
, Kim%
\BCBL {}\ \BBA {} Lee%
}{%
J.~Kim%
\ \protect \BOthers {.}}{%
{\protect \APACyear {2023}}%
}]{%
Kim2023twod}
\APACinsertmetastar {%
Kim2023twod}%
\begin{APACrefauthors}%
Kim, J.%
, Kim, J.%
\BCBL {}\ \BBA {} Lee, C.%
\end{APACrefauthors}%
\unskip\
\newblock
\APACrefYearMonthDay{2023}{}{}.
\newblock
{\BBOQ}\APACrefatitle {Prediction and control of two-dimensional decaying
  turbulence using generative adversarial networks} {Prediction and control of
  two-dimensional decaying turbulence using generative adversarial
  networks}.{\BBCQ}
\newblock
\APACjournalVolNumPages{arXiv preprint arXIv:2312.07037}{}{}{}.
\PrintBackRefs{\CurrentBib}

\bibitem [\protect \citeauthoryear {%
J.~Kim%
\ \BBA {} Lee%
}{%
J.~Kim%
\ \BBA {} Lee%
}{%
{\protect \APACyear {2020}}%
{\protect \APACexlab {{\protect \BCnt {1}}}}}]{%
Kim2020inflow}
\APACinsertmetastar {%
Kim2020inflow}%
\begin{APACrefauthors}%
Kim, J.%
\BCBT {}\ \BBA {} Lee, C.%
\end{APACrefauthors}%
\unskip\
\newblock
\APACrefYearMonthDay{2020{\protect \BCnt {1}}}{}{}.
\newblock
{\BBOQ}\APACrefatitle {Deep unsupervised learning of turbulence for inflow
  generation at various Reynolds numbers} {Deep unsupervised learning of
  turbulence for inflow generation at various reynolds numbers}.{\BBCQ}
\newblock
\APACjournalVolNumPages{Journal of Computational Physics}{406}{}{}.
\newblock
\begin{APACrefDOI} \doi{10.1016/j.jcp.2019.109216} \end{APACrefDOI}
\PrintBackRefs{\CurrentBib}

\bibitem [\protect \citeauthoryear {%
J.~Kim%
\ \BBA {} Lee%
}{%
J.~Kim%
\ \BBA {} Lee%
}{%
{\protect \APACyear {2020}}%
{\protect \APACexlab {{\protect \BCnt {2}}}}}]{%
Kim2020heat}
\APACinsertmetastar {%
Kim2020heat}%
\begin{APACrefauthors}%
Kim, J.%
\BCBT {}\ \BBA {} Lee, C.%
\end{APACrefauthors}%
\unskip\
\newblock
\APACrefYearMonthDay{2020{\protect \BCnt {2}}}{}{}.
\newblock
{\BBOQ}\APACrefatitle {Prediction of turbulent heat transfer using
  convolutional neural networks} {Prediction of turbulent heat transfer using
  convolutional neural networks}.{\BBCQ}
\newblock
\APACjournalVolNumPages{Journal of Fluid Mechanics}{882}{}{}.
\newblock
\begin{APACrefDOI} \doi{10.1017/jfm.2019.814} \end{APACrefDOI}
\PrintBackRefs{\CurrentBib}

\bibitem [\protect \citeauthoryear {%
{Korea Meteorological Administration}%
}{%
{Korea Meteorological Administration}%
}{%
{\protect \APACyear {2021}}%
}]{%
kmaAlmanac}
\APACinsertmetastar {%
kmaAlmanac}%
\begin{APACrefauthors}%
{Korea Meteorological Administration}.%
\end{APACrefauthors}%
\unskip\
\newblock
\APACrefYear{2021}.
\newblock
\APACrefbtitle {{2020 Meteorological Almanac}} {{2020 Meteorological Almanac}}.
\newblock
\APACaddressPublisher{}{Korea Meteorological Administration}.
\PrintBackRefs{\CurrentBib}

\bibitem [\protect \citeauthoryear {%
Lam%
\ \protect \BOthers {.}}{%
Lam%
\ \protect \BOthers {.}}{%
{\protect \APACyear {2023}}%
}]{%
lam2022}
\APACinsertmetastar {%
lam2022}%
\begin{APACrefauthors}%
Lam, R.%
, Sanchez-Gonzalez, A.%
, Willson, M.%
, Wirnsberger, P.%
, Fortunato, M.%
, Pritzel, A.%
\BDBL {}others%
\end{APACrefauthors}%
\unskip\
\newblock
\APACrefYearMonthDay{2023}{}{}.
\newblock
{\BBOQ}\APACrefatitle {GraphCast: Learning skillful medium-range global weather
  forecasting} {Graphcast: Learning skillful medium-range global weather
  forecasting}.{\BBCQ}
\newblock
\APACjournalVolNumPages{Science}{}{}{10.1126/science.adi2336}.
\PrintBackRefs{\CurrentBib}

\bibitem [\protect \citeauthoryear {%
Lee%
, Kim%
\BCBL {}\ \BBA {} Lee%
}{%
Lee%
\ \protect \BOthers {.}}{%
{\protect \APACyear {2023}}%
}]{%
Lee2023drag}
\APACinsertmetastar {%
Lee2023drag}%
\begin{APACrefauthors}%
Lee, T.%
, Kim, J.%
\BCBL {}\ \BBA {} Lee, C.%
\end{APACrefauthors}%
\unskip\
\newblock
\APACrefYearMonthDay{2023}{}{}.
\newblock
{\BBOQ}\APACrefatitle {Turbulence control for drag reduction through deep
  reinforcement learning} {Turbulence control for drag reduction through deep
  reinforcement learning}.{\BBCQ}
\newblock
\APACjournalVolNumPages{Physical Review Fluids}{8}{}{024604}.
\PrintBackRefs{\CurrentBib}

\bibitem [\protect \citeauthoryear {%
Liu%
\ \protect \BOthers {.}}{%
Liu%
\ \protect \BOthers {.}}{%
{\protect \APACyear {2022}}%
}]{%
liu2022swin}
\APACinsertmetastar {%
liu2022swin}%
\begin{APACrefauthors}%
Liu, Z.%
, Hu, H.%
, Lin, Y.%
, Yao, Z.%
, Xie, Z.%
, Wei, Y.%
\BDBL {}others%
\end{APACrefauthors}%
\unskip\
\newblock
\APACrefYearMonthDay{2022}{}{}.
\newblock
{\BBOQ}\APACrefatitle {Swin transformer v2: Scaling up capacity and resolution}
  {Swin transformer v2: Scaling up capacity and resolution}.{\BBCQ}
\newblock
\BIn{} \APACrefbtitle {{Proceedings of the IEEE/CVF Conference on Computer
  Vision and Pattern Recognition}} {{Proceedings of the IEEE/CVF Conference on
  Computer Vision and Pattern Recognition}}\ (\BPGS\ 12009--12019).
\PrintBackRefs{\CurrentBib}

\bibitem [\protect \citeauthoryear {%
Liu%
\ \protect \BOthers {.}}{%
Liu%
\ \protect \BOthers {.}}{%
{\protect \APACyear {2021}}%
}]{%
Liu_2021_ICCV}
\APACinsertmetastar {%
Liu_2021_ICCV}%
\begin{APACrefauthors}%
Liu, Z.%
, Lin, Y.%
, Cao, Y.%
, Hu, H.%
, Wei, Y.%
, Zhang, Z.%
\BDBL {}Guo, B.%
\end{APACrefauthors}%
\unskip\
\newblock
\APACrefYearMonthDay{2021}{}{}.
\newblock
{\BBOQ}\APACrefatitle {Swin Transformer: Hierarchical Vision Transformer Using
  Shifted Windows} {Swin transformer: Hierarchical vision transformer using
  shifted windows}.{\BBCQ}
\newblock
\BIn{} \APACrefbtitle {{Proceedings of the IEEE/CVF International Conference on
  Computer Vision (ICCV)}} {{Proceedings of the IEEE/CVF International
  Conference on Computer Vision (ICCV)}}\ (\BPG~10012-10022).
\PrintBackRefs{\CurrentBib}

\bibitem [\protect \citeauthoryear {%
Mooney%
, Mulligan%
\BCBL {}\ \BBA {} Fealy%
}{%
Mooney%
\ \protect \BOthers {.}}{%
{\protect \APACyear {2011}}%
}]{%
mooney2011}
\APACinsertmetastar {%
mooney2011}%
\begin{APACrefauthors}%
Mooney, P\BPBI A.%
, Mulligan, F\BPBI J.%
\BCBL {}\ \BBA {} Fealy, R.%
\end{APACrefauthors}%
\unskip\
\newblock
\APACrefYearMonthDay{2011}{}{}.
\newblock
{\BBOQ}\APACrefatitle {Comparison of ERA-40, ERA-Interim and NCEP/NCAR
  reanalysis data with observed surface air temperatures over Ireland}
  {Comparison of era-40, era-interim and ncep/ncar reanalysis data with
  observed surface air temperatures over ireland}.{\BBCQ}
\newblock
\APACjournalVolNumPages{International Journal of Climatology}{31}{4}{545--557}.
\PrintBackRefs{\CurrentBib}

\bibitem [\protect \citeauthoryear {%
Nakkiran%
\ \protect \BOthers {.}}{%
Nakkiran%
\ \protect \BOthers {.}}{%
{\protect \APACyear {2021}}%
}]{%
nakkiran2021deep}
\APACinsertmetastar {%
nakkiran2021deep}%
\begin{APACrefauthors}%
Nakkiran, P.%
, Kaplun, G.%
, Bansal, Y.%
, Yang, T.%
, Barak, B.%
\BCBL {}\ \BBA {} Sutskever, I.%
\end{APACrefauthors}%
\unskip\
\newblock
\APACrefYearMonthDay{2021}{}{}.
\newblock
{\BBOQ}\APACrefatitle {Deep double descent: Where bigger models and more data
  hurt} {Deep double descent: Where bigger models and more data hurt}.{\BBCQ}
\newblock
\APACjournalVolNumPages{Journal of Statistical Mechanics: Theory and
  Experiment}{2021}{12}{124003}.
\PrintBackRefs{\CurrentBib}

\bibitem [\protect \citeauthoryear {%
Park%
\ \BBA {} Lee%
}{%
Park%
\ \BBA {} Lee%
}{%
{\protect \APACyear {2023}}%
{\protect \APACexlab {{\protect \BCnt {1}}}}}]{%
park_2023_10252672}
\APACinsertmetastar {%
park_2023_10252672}%
\begin{APACrefauthors}%
Park, J.%
\BCBT {}\ \BBA {} Lee, C.%
\end{APACrefauthors}%
\unskip\
\newblock
\APACrefYearMonthDay{2023{\protect \BCnt {1}}}{}{}.
\newblock
\APACrefbtitle {{The codes for TPTNet}} {{The codes for TPTNet}}\ [Software].
\newblock
\APACaddressPublisher{}{Zenodo}.
\newblock
\begin{APACrefURL} \url{https://doi.org/10.5281/zenodo.10252672}
  \end{APACrefURL}
\newblock
\begin{APACrefDOI} \doi{10.5281/zenodo.10252672} \end{APACrefDOI}
\PrintBackRefs{\CurrentBib}

\bibitem [\protect \citeauthoryear {%
Park%
\ \BBA {} Lee%
}{%
Park%
\ \BBA {} Lee%
}{%
{\protect \APACyear {2023}}%
{\protect \APACexlab {{\protect \BCnt {2}}}}}]{%
DVN/J3DRDT_2023}
\APACinsertmetastar {%
DVN/J3DRDT_2023}%
\begin{APACrefauthors}%
Park, J.%
\BCBT {}\ \BBA {} Lee, C.%
\end{APACrefauthors}%
\unskip\
\newblock
\APACrefYearMonthDay{2023{\protect \BCnt {2}}}{}{}.
\newblock
\APACrefbtitle {{Datasets for TPTNet}} {{Datasets for TPTNet}}\ [Dataset].
\newblock
\APACaddressPublisher{}{Harvard Dataverse}.
\newblock
\begin{APACrefURL} \url{https://doi.org/10.7910/DVN/J3DRDT} \end{APACrefURL}
\newblock
\begin{APACrefDOI} \doi{10.7910/DVN/J3DRDT} \end{APACrefDOI}
\PrintBackRefs{\CurrentBib}

\bibitem [\protect \citeauthoryear {%
Pathak%
\ \protect \BOthers {.}}{%
Pathak%
\ \protect \BOthers {.}}{%
{\protect \APACyear {2022}}%
}]{%
Pathak2022}
\APACinsertmetastar {%
Pathak2022}%
\begin{APACrefauthors}%
Pathak, J.%
, Subramanian, S.%
, Harrington, P.%
, Raja, S.%
, Chattopadhyay, A.%
, Mardani, M.%
\BDBL {}others%
\end{APACrefauthors}%
\unskip\
\newblock
\APACrefYearMonthDay{2022}{}{}.
\newblock
{\BBOQ}\APACrefatitle {{Fourcastnet: A global data-driven high-resolution
  weather model using adaptive Fourier neural operators}} {{Fourcastnet: A
  global data-driven high-resolution weather model using adaptive Fourier
  neural operators}}.{\BBCQ}
\newblock
\APACjournalVolNumPages{arXiv preprint arXiv:2202.11214}{}{}{}.
\PrintBackRefs{\CurrentBib}

\bibitem [\protect \citeauthoryear {%
Ramavajjala%
\ \BBA {} Mitra%
}{%
Ramavajjala%
\ \BBA {} Mitra%
}{%
{\protect \APACyear {2023}}%
}]{%
ramavajjala2023verification}
\APACinsertmetastar {%
ramavajjala2023verification}%
\begin{APACrefauthors}%
Ramavajjala, V.%
\BCBT {}\ \BBA {} Mitra, P\BPBI P.%
\end{APACrefauthors}%
\unskip\
\newblock
\APACrefYearMonthDay{2023}{}{}.
\newblock
{\BBOQ}\APACrefatitle {Verification against in-situ observations for
  Data-Driven Weather Prediction} {Verification against in-situ observations
  for data-driven weather prediction}.{\BBCQ}
\newblock
\APACjournalVolNumPages{arXiv preprint arXiv:2305.00048}{}{}{}.
\PrintBackRefs{\CurrentBib}

\bibitem [\protect \citeauthoryear {%
Rasp%
\ \protect \BOthers {.}}{%
Rasp%
\ \protect \BOthers {.}}{%
{\protect \APACyear {2020}}%
}]{%
Rasp2020}
\APACinsertmetastar {%
Rasp2020}%
\begin{APACrefauthors}%
Rasp, S.%
, Dueben, P\BPBI D.%
, Scher, S.%
, Weyn, J\BPBI A.%
, Mouatadid, S.%
\BCBL {}\ \BBA {} Thuerey, N.%
\end{APACrefauthors}%
\unskip\
\newblock
\APACrefYearMonthDay{2020}{}{}.
\newblock
{\BBOQ}\APACrefatitle {WeatherBench: A Benchmark Data Set for Data-Driven
  Weather Forecasting} {Weatherbench: A benchmark data set for data-driven
  weather forecasting}.{\BBCQ}
\newblock
\APACjournalVolNumPages{Journal of Advances in Modeling Earth Systems}{12}{}{}.
\newblock
\begin{APACrefDOI} \doi{10.1029/2020MS002203} \end{APACrefDOI}
\PrintBackRefs{\CurrentBib}

\bibitem [\protect \citeauthoryear {%
Rasp%
\ \BBA {} Thuerey%
}{%
Rasp%
\ \BBA {} Thuerey%
}{%
{\protect \APACyear {2021}}%
}]{%
Rasp2021}
\APACinsertmetastar {%
Rasp2021}%
\begin{APACrefauthors}%
Rasp, S.%
\BCBT {}\ \BBA {} Thuerey, N.%
\end{APACrefauthors}%
\unskip\
\newblock
\APACrefYearMonthDay{2021}{}{}.
\newblock
{\BBOQ}\APACrefatitle {Data-Driven Medium-Range Weather Prediction With a
  Resnet Pretrained on Climate Simulations: A New Model for WeatherBench}
  {Data-driven medium-range weather prediction with a resnet pretrained on
  climate simulations: A new model for weatherbench}.{\BBCQ}
\newblock
\APACjournalVolNumPages{Journal of Advances in Modeling Earth Systems}{13}{}{}.
\newblock
\begin{APACrefDOI} \doi{10.1029/2020MS002405} \end{APACrefDOI}
\PrintBackRefs{\CurrentBib}

\bibitem [\protect \citeauthoryear {%
Schultz%
\ \protect \BOthers {.}}{%
Schultz%
\ \protect \BOthers {.}}{%
{\protect \APACyear {2021}}%
}]{%
Schultz2021}
\APACinsertmetastar {%
Schultz2021}%
\begin{APACrefauthors}%
Schultz, M\BPBI G.%
, Betancourt, C.%
, Gong, B.%
, Kleinert, F.%
, Langguth, M.%
, Leufen, L\BPBI H.%
\BDBL {}Stadtler, S.%
\end{APACrefauthors}%
\unskip\
\newblock
\APACrefYearMonthDay{2021}{}{}.
\newblock
{\BBOQ}\APACrefatitle {Can deep learning beat numerical weather prediction?}
  {Can deep learning beat numerical weather prediction?}{\BBCQ}
\newblock
\APACjournalVolNumPages{Philosophical Transactions of the Royal Society
  A}{379}{2194}{20200097}.
\PrintBackRefs{\CurrentBib}

\bibitem [\protect \citeauthoryear {%
Veli{\v{c}}kovi{\'c}%
\ \protect \BOthers {.}}{%
Veli{\v{c}}kovi{\'c}%
\ \protect \BOthers {.}}{%
{\protect \APACyear {2017}}%
}]{%
velivckovic2017graph}
\APACinsertmetastar {%
velivckovic2017graph}%
\begin{APACrefauthors}%
Veli{\v{c}}kovi{\'c}, P.%
, Cucurull, G.%
, Casanova, A.%
, Romero, A.%
, Lio, P.%
\BCBL {}\ \BBA {} Bengio, Y.%
\end{APACrefauthors}%
\unskip\
\newblock
\APACrefYearMonthDay{2017}{}{}.
\newblock
{\BBOQ}\APACrefatitle {Graph attention networks} {Graph attention
  networks}.{\BBCQ}
\newblock
\APACjournalVolNumPages{arXiv preprint arXiv:1710.10903}{}{}{}.
\PrintBackRefs{\CurrentBib}

\bibitem [\protect \citeauthoryear {%
Weyn%
, Durran%
, Caruana%
\BCBL {}\ \BBA {} Cresswell-Clay%
}{%
Weyn%
\ \protect \BOthers {.}}{%
{\protect \APACyear {2021}}%
}]{%
Weyn2021}
\APACinsertmetastar {%
Weyn2021}%
\begin{APACrefauthors}%
Weyn, J\BPBI A.%
, Durran, D\BPBI R.%
, Caruana, R.%
\BCBL {}\ \BBA {} Cresswell-Clay, N.%
\end{APACrefauthors}%
\unskip\
\newblock
\APACrefYearMonthDay{2021}{}{}.
\newblock
{\BBOQ}\APACrefatitle {Sub-Seasonal Forecasting With a Large Ensemble of
  Deep-Learning Weather Prediction Models} {Sub-seasonal forecasting with a
  large ensemble of deep-learning weather prediction models}.{\BBCQ}
\newblock
\APACjournalVolNumPages{Journal of Advances in Modeling Earth Systems}{13}{}{}.
\newblock
\begin{APACrefDOI} \doi{10.1029/2021MS002502} \end{APACrefDOI}
\PrintBackRefs{\CurrentBib}

\bibitem [\protect \citeauthoryear {%
{World Meteorological Organization}%
}{%
{World Meteorological Organization}%
}{%
{\protect \APACyear {2021}}%
}]{%
WMO-Guidelines}
\APACinsertmetastar {%
WMO-Guidelines}%
\begin{APACrefauthors}%
{World Meteorological Organization}.%
\end{APACrefauthors}%
\unskip\
\newblock
\APACrefYear{2021}.
\newblock
\APACrefbtitle {{Guidelines on Surface Station Data Quality Control and Quality
  Assurance for Climate Applications}} {{Guidelines on Surface Station Data
  Quality Control and Quality Assurance for Climate Applications}}.
\newblock
\APACaddressPublisher{}{World Meteorological Organization}.
\PrintBackRefs{\CurrentBib}

\bibitem [\protect \citeauthoryear {%
Zhu%
\ \protect \BOthers {.}}{%
Zhu%
\ \protect \BOthers {.}}{%
{\protect \APACyear {2023}}%
}]{%
Zhu2023}
\APACinsertmetastar {%
Zhu2023}%
\begin{APACrefauthors}%
Zhu, X.%
, Xiong, Y.%
, Wu, M.%
, Nie, G.%
, Zhang, B.%
\BCBL {}\ \BBA {} Yang, Z.%
\end{APACrefauthors}%
\unskip\
\newblock
\APACrefYearMonthDay{2023}{}{}.
\newblock
{\BBOQ}\APACrefatitle {Weather2K: A Multivariate Spatio-Temporal Benchmark
  Dataset for Meteorological Forecasting Based on Real-Time Observation Data
  from Ground Weather Stations} {Weather2k: A multivariate spatio-temporal
  benchmark dataset for meteorological forecasting based on real-time
  observation data from ground weather stations}.{\BBCQ}
\newblock
\APACjournalVolNumPages{arXiv preprint arXiv:2302.10493}{}{}{}.
\PrintBackRefs{\CurrentBib}

\end{thebibliography}

%
%
%
%
%

\end{document}